\documentclass{article}
\usepackage{microtype}
\usepackage{graphicx}
\usepackage{wrapfig}
\usepackage{booktabs} 
\usepackage{makecell}
\usepackage{graphicx}
\usepackage{caption}
\usepackage{courier}
\usepackage{multirow}
\usepackage{color}
\usepackage{adjustbox}
\usepackage{subcaption}
\usepackage{amssymb,amsthm}
\usepackage{amsmath}
\usepackage{gensymb}
\usepackage{amsfonts}       
\usepackage{pifont}
\usepackage{enumitem}
\usepackage{multirow}
\newcommand{\xhdr}[1]{{\noindent\bfseries #1}.}
\usepackage{algorithm}  
\usepackage[noend]{algorithmic}
\usepackage{bbding}
\usepackage{pifont}
\usepackage{wasysym}
\usepackage{amssymb}
\usepackage{amsmath,nccmath}
\usepackage{amsmath,amsfonts,amssymb,bbm, bm}
\usepackage[subtle]{savetrees}
\definecolor{mydarkblue}{rgb}{0,0.08,0.45}
\definecolor{myorange}{RGB}{2, 142, 2}
\usepackage{hyperref}
\hypersetup{
    colorlinks=true,
    linkcolor=mydarkblue,
    filecolor=magenta,      
    urlcolor=cyan,
    citecolor=teal,
    citecolor=myorange,
}

\usepackage{bbm}
\usepackage[accepted, nohyperref]{icml2023}

\usepackage{url}

\newcommand{\rebuttal}[1]{\textcolor{black}{#1}}

\newcommand{\ie}{\textit{i.e., }}
\newcommand{\eg}{\textit{e.g., }}

\newcommand{\etc}{\textit{etc.}}
\newcommand{\wrt}{\textit{w.r.t. }}

\newcommand{\Prob}{\mathbb{P}}

\newtheorem{theorem}{Theorem}
\newtheorem{definition}{Definition}
\newtheorem{observation}{Observation}

\usepackage{pifont}
\newcommand{\xmark}{\ding{55}}%

\newtheorem*{theorem*}{Theorem}

\usepackage{color, colortbl}
\definecolor{codegreen}{rgb}{0,0.6,0}
\definecolor{codeblue}{rgb}{0,0,0.6}
\definecolor{codegray}{rgb}{0.5,0.5,0.5}
\definecolor{codepurple}{rgb}{0.58,0,0.82}
\definecolor{backcolour}{rgb}{0.95,0.95,0.92}
\definecolor{lightblue}{HTML}{84C7F9}
\definecolor{lighterblue}{HTML}{D4ECFF}
\definecolor{myblue}{HTML}{D4ECFF}
\definecolor{mygreen}{HTML}{D0F0C0}
\definecolor{mycham}{HTML}{F7E7CE}
\definecolor{mygray}{gray}{0.90}
\newcommand{\ours}{DISC }
\newcommand{\ourst}{DISC}

\def\R{\mathbbm{R}}

\def\E{\mathbbm{E}}

\usepackage{tabularx}

\begin{document}
\twocolumn[
\icmltitle{
Discover and Cure: 
Concept-aware Mitigation of Spurious Correlation}
\icmlsetsymbol{equal}{*}

\begin{icmlauthorlist}
\icmlauthor{Shirley Wu}{stanford}
\icmlauthor{Mert Yuksekgonul}{stanford}
\icmlauthor{Linjun Zhang}{rugters}
\icmlauthor{James Zou}{stanford}
\end{icmlauthorlist}

\icmlaffiliation{stanford}{Department of Computer Science, Stanford University.}
\icmlaffiliation{rugters}{Department of Statistics, Rutgers University}

\icmlcorrespondingauthor{Shirley Wu}{shirwu@cs.stanford.edu}
\icmlcorrespondingauthor{James Zou}{jamesz@stanford.edu}

\icmlkeywords{Machine Learning, ICML}

\vskip 0.3in

]

\printAffiliationsAndNotice{}

\begin{abstract}

Deep neural networks often rely on spurious correlations to make predictions, which hinders generalization beyond training environments. For instance, models that associate cats with bed backgrounds can fail to predict the existence of cats in other environments without beds. 
Mitigating spurious correlations is crucial in building trustworthy models. However, the existing works lack transparency to offer insights into the mitigation process.
In this work, we propose an interpretable framework, \underline{Dis}cover and \underline{C}ure (\ourst), to tackle the issue. With human-interpretable concepts, \ours iteratively 1) discovers unstable concepts across different environments as spurious attributes, then 2) intervenes on the training data using the discovered concepts to reduce spurious correlation. Across systematic experiments, \ours provides superior generalization ability and interpretability than the existing approaches. Specifically, it outperforms the state-of-the-art methods on an object recognition task and a skin-lesion classification task by 7.5\% and 9.6\%, respectively. 
Additionally, we offer theoretical analysis and guarantees to understand the benefits of models trained by DISC. Code and data are available at \href{https://github.com/Wuyxin/DISC}{https://github.com/Wuyxin/DISC}.


\end{abstract}
\section{Introduction}
\begin{figure}[t]
    \centering     
    \includegraphics[width=0.45\textwidth]{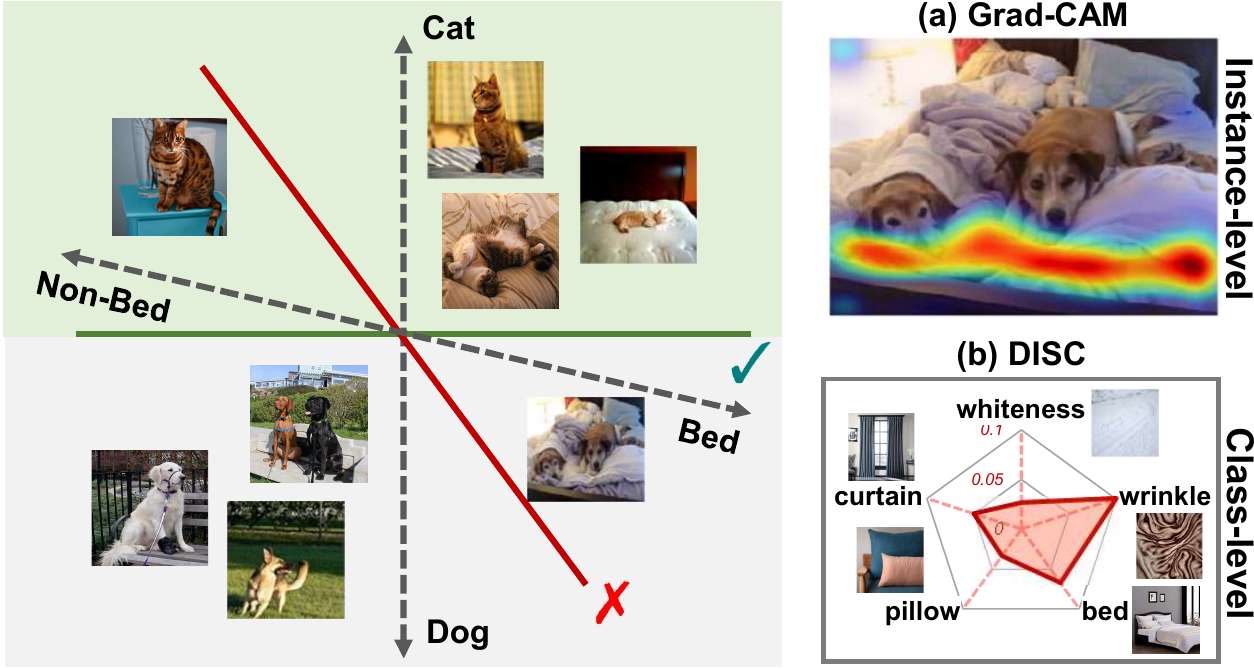}
    \vspace{-5pt}
    \caption{Left: The dog/cat classifiers that rely (red) or do not rely (green) on spurious correlations; Right: Spuriousness discovery results of Grad-CAM~\cite{Grad-CAM} and \ourst, where we propose a class-level metric to indicate the degree of spurious correlation between concepts and the ``cat'' class.
}   
    \label{fig:example}
\end{figure}

Spurious correlations are common in real-world data analysis. 
Spurious attributes are typically associated with the class label but are non-generalizable~\cite{KaushikHL20, groupdro}.
For example, as shown in Figure~\ref{fig:example}, neural networks that mistakenly associate cats with beds are prone to fail in different settings, \eg dog-on-bed or cat-on-desk, where the spurious correlation no longer holds.
This lack of reliability is a central issue in critical applications, \eg medical diagnosis~\cite{isic-split}. 

Existing works have developed methods to mitigate the spurious correlations inside deep models. 
For instance, invariant learning~\cite{irm, ibirm} learns a stable representation across environments to avoid varying factors, including spurious attributes. 
Leveraging the vulnerability of Empirical Risk Minimization (ERM) models towards spurious attributes, some works upweights over-confident~\cite{lff} or misclassified~\cite{jtt} instances from a trained ERM model to counteract the spurious correlation.

However, these works 
lack interpretability into what are the information the model is learning or ignoring, which hinders human understanding and model auditing.
While post-hoc explainability methods~\cite{Grad-CAM, lime, shap} 
offer visualization that could contain spurious regions, it is still ambiguous to understand.
For example, in Figure~\ref{fig:example} (a), the highlighted region shows the attributes that contribute most to the prediction, explaining why the image is mistakenly predicted as ``cat'' but ``dog''. Nevertheless, it is not clear which of the attributes (\eg whiteness, wrinkle texture, or items like pillow) mostly contributes to the spurious correlation. Moreover, such instance-level interpretations are not informative about the overall spurious correlations existing in the class.

In this work, we adopt concepts that align with human understandings to discover class-level spurious attributes, leveraging a concept bank as an auxiliary knowledge base.
We show that the invariant concepts, \eg the shape of cats, remain stable across data environments, while the spurious concepts, \eg ``bed'', have inconstant existence across the instances within the
class. Inspired by this property, we propose a class-level metric, \textit{concept sensitivity}, to quantify a concept's instability across the data environments. For example, in Figure~\ref{fig:example} (b), we identify both ``wrinkle'' and ``bed'' as  highly spurious concepts of ``cat'' based on the large magnitude of concept sensitivity, and we
refer to this step as \textbf{the discovery step}. 


Upon discovering the spurious concepts, we propose an intervention step, namely \emph{concept-aware intervention}, to reduce the models'
reliance on spurious concepts.
The high-level idea is to intervene on the selected classes with spurious concepts to maintain a balanced distribution of the spurious concepts. 
For instance, after identifying that ``wrinkle'' and ``bed'' are spurious concepts correlated with the ``cat'' class, we use concept images of them to intervene in the ``dog'' class, as shown in the bottom right of Figure~\ref{fig:framework}. 
With a balanced distribution of spurious concepts across different classes, we prevent the model from taking advantage of spurious concepts to make predictions, thus canceling the spurious bias. 
We refer to this process as \textbf{the cure step}. 

\xhdr{Discover and Cure} 
Finally, our algorithm, \ourst, iteratively conducts the \underline{dis}covery and \underline{c}ure steps during training. In each iteration, it discovers the spurious concepts for the current model. Then based on the discovered concepts, it intervenes on the training datasets to remove the spurious correlations, on top of which the model is updated. 
Here we focus on image classification tasks. Empirically, \ours discovers spurious concepts that align with ground truth spurious attributes and outperforms the state-of-the-art baselines averagely with a large margin. 
Our \textbf{contributions} are:
\vspace{-10pt}
\begin{itemize}[leftmargin=*]
    \item We develop a novel and interpretable framework to discover spurious concepts and effectively mitigate spurious correlations for model generalization. 
\vspace{-5pt}
    \item We empirically validate our method's effectiveness on diverse datasets and reveal insights into how models overcome spurious correlations.
\vspace{-5pt}
    \item We theoretically guarantee the convergence and generalization ability of the models trained by \ourst.
\end{itemize}
\vspace{-10pt}


\section{Related Work}
Our work, built on human-interpretable concepts, involves discovering and curing spurious correlations. Here we discuss three classes of related works:


\xhdr{Concepts}
Concepts, \eg \textit{blueness} or \textit{stripes}, are human-interpretable semantics. Concepts have been used to build interpretable models~\cite{LampertNH09, KumarBBN09, cbm, pcbm}, or used in a post-hoc manner~\cite{network-dissection, tcav} to interpret the predictions of deep neural networks (DNNs).
Specifically, \citet{tcav} introduce Concept Activation Vectors~(CAVs), where a CAV represents the direction in the hidden space of a DNN that corresponds to the existence of a concept, helping align the internal state of DNNs with human expectations.

\xhdr{Discovering Spurious Correlations} Previous works study spurious correlations in settings like image texture and backgrounds~\cite{GeirhosRMBWB19, groupdro}, domain shifts~\cite{wilds, domainbed, breeds,ye2023freeze}, and causally unstable attributes~\cite{irm, dir}. 
Detecting spurious correlations reveals model biases that are harmful to generalization. 
Some works obtain spurious attributes using domain knowledge~\cite{ClarkYZ19,KaushikHL20,Nauta21}, however, spurious attributes could go beyond domain knowledge.
For instance, \citet{eiil} infer spurious attributes by learning environments.
\citet{george, bias-attribute} cluster a model’s embeddings and use the clusters to reveal spurious attributes. 

Recent works~\cite{abs-2106, abs-2211, cce} also use explainability techniques to find spurious attributes and require human inspection. 
Unlike instance-level auditing, we propose a class-level metric 
which offers high-level interpretability that is more reliable and user-friendly.
\rebuttal{Moreover, concept-level and interactive debugging methods~\cite{partprot, debias_concept, exp_inter} leverage concepts or human feedback to perform debugging. See \citet{exp_inter_overview} for an overview. For example, \citet{partprot} propose ProtoPDebug that allows a human supervisor to provide feedback to part-prototype networks~\cite{ChenLTBRS19} (ProtoPNets) on the model's explanations. In contrast to our method, they generally work with a restricted class of models (\eg CBMs~\cite{cbm} or ProtoPNets) and often require human annotation to identify the concepts. See Table~\ref{tab:comparison} for the comparison between the selected works and our method.}





\vspace{-3pt}

\xhdr{Curing Spurious Bias} 
Learning spurious attributes makes models over-sensitive to spurious factors and their distribution shifts, which is related to invariant and robust learning.
\begin{itemize}[leftmargin=*]
\item \vspace{-5pt}
    \textbf{Invariant Learning}. 
    \citet{irm} propose learning an invariant encoder such that the downstream classifiers are optimal in different environments. 
    Other works target invariance via correlation alignment~\cite{coral}, variance penalty~\cite{rex, unshuffling}, and gradient alignment~\cite{fish} across domains. However, these are not interpretable, which provides little insight into the data bias.
\item \vspace{-5pt}
    \textbf{Instance Reweighting}. 
    Instance reweighting puts high importance on examples that unlikely contain spurious attributes to remove bias~\cite{YaghoobzadehMCH21, UtamaMG20, Dagaev21, ZhangSZFR22, lff, DebiAN}.
    Despite its simplicity, such instances could be rare when models perfectly fit the training data, which limits the effectiveness. 
    Distributionally Robust Optimizaton (DRO)~\cite{Ben-TalHWMR13, OrenSHL19, groupdro, zhang2020coping} is a special case that puts more weights on observations with high loss~\cite{NamkoongD16, HuNSS18, LevyCDS20}. 
    Yet, the impact of instance reweighting on over-parameterized DNNs could diminish over epochs~\cite{ByrdL19}, leading to overfitting eventually. 
\vspace{-6pt}
\item \textbf{Data Augmentation}. 
    Other works use data augmentations like adversarial mixup \cite{dm-ada}, selective augmentation~\cite{lisa}, and uncertainty-aware mixup~\cite{umix} to reduce the reliance on the spurious correlation~\cite{zhang2020does}. 
    \rebuttal{Moreover, \citet{RegMixup} propose that mixup~\cite{mixup} as a regularizer can further improves out-of-distribution robustness. }
    However, these augmentations do not explicitly consider multiple and coexistent spurious attributes, which is common in real-world applications. 
    With a concept bank generated from a text-to-image generator, \ours detects the spurious concepts and adaptively mixes up concept images with instances in selected classes. 
    Concurrent work~\cite{distill-failure} uses a captioning model to capture the failure mode and generate synthetic images for fine-tuning. Nevertheless, the generated data relies on the captioning model, which can be out-of-distribution and infeasible for hard-to-describe datasets like skin lesion images.
    \ours is a more flexible solution using concept images to do the intervention. 
\end{itemize}

\section{Method}
\label{sec:method}
Here, we describe the problem setup and our method. We formalize our problem setup in Section~\ref{sec:fomulation} and introduce the construction of the concept bank in Section~\ref{sec:concept_bank}. 
Then, we discuss the discovery of spurious concepts in Section~\ref{sec:search_shortcut} and the removal of spurious correlations in Section~\ref{sec:reduce_sens}. For clarity, we summarize the \underline{main notation} in Appendix~\ref{app:notation}.

\subsection{Problem Formulation}
\label{sec:fomulation}

We consider a supervised image classification problem.
Specifically, we are given a training dataset $\mathcal{D}_{tr}=\{(x_1, y_1),\ldots, (x_n, y_n)\}$. We define $\mathcal{Y}$ as the label space and $\mathcal{P}_{tr}$ as the distribution of the training dataset.



For an arbitrary loss function $\ell$, Empirical Risk Minimization (ERM) minimizes the empirical loss for a model $f$:
\vspace{-5pt}
\begin{equation}
\arg \min _{\phi} \mathbb{E}_{(x, y) \sim \mathcal{D}_{tr}}\left[\ell\left(f_\phi(x), y\right)\right]
\label{eq:erm}
\vspace{-10pt}
\end{equation}

where $f$ is parameterized by $\phi$. 
Due to the unstable nature of spurious attributes, the test distribution $\mathcal{P}_{te}$ is often different from the training distribution, \ie $\mathcal{P}_{te}\neq \mathcal{P}_{tr}$. Thus, the model trained with the ERM falls short of generalizing to datasets $\mathcal{D}_{te}\sim \mathcal{P}_{te}$ where the spurious correlations shift or no longer hold.
Thus, our goal is to overcome the model's bias in the presence of spurious correlations. From a causality perspective, spurious attributes are defined as the attributes $F$ that are not causally related to the truth label $Y$, but are correlated with the truth label $Y$ in the training data due to data sampling bias or imbalance.  
For example, ``bed'' can not determine the image being labeled as ``cat'', but may be correlated to the label ``cat'' if the cat images are mostly taken in bedrooms. For our purpose, we consider spurious attributes to be attributes whose presence is correlated with the label in some environments but not others.




\begin{figure*}
    \centering    
    \includegraphics[width=0.80\textwidth]{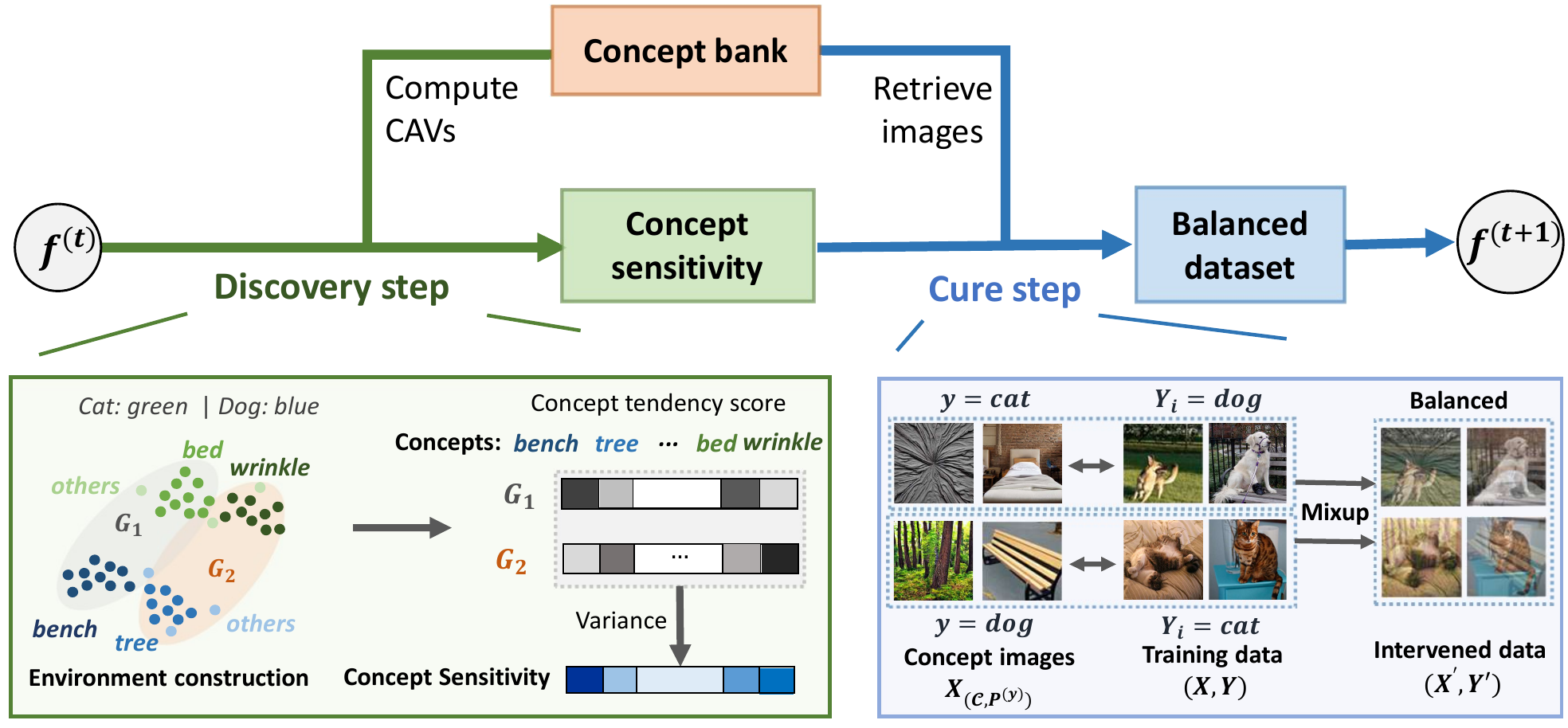}
    \vspace{-6pt}
    \caption{\textbf{\ours Framework.} 
    At $t$-th iteration, \ours computes the concept sensitivities based on the previously constructed environments and the CAVs from the concept bank, which discovers \textit{(wrinkle, bed)} and \textit{(bench, tree)} as the spurious concepts of ``cat'' and ``dog'',  respectively. In the cure step, based on the concept sensitivities, DISC retrieves concept images, \eg ``bed'', from the concept bank and mixes them up with the training subset, \eg dog images, to remove the spurious correlation. Finally, the model is updated on the balanced dataset.
    }
    \vspace{-10pt}
    \label{fig:framework}
\end{figure*}

\subsection{Concept Bank}
\label{sec:concept_bank}
To describe the spurious attributes, we consider them as concepts in a human-understandable fashion instead of pixel-level patterns. 
We build a comprehensive concept bank that widely covers potential spurious concept candidates. Formally, we have


\begin{definition}[\textbf{Concept bank}]
    A concept bank with $m$ concepts can be expressed as $\mathcal{C}:=\{(c_i, \mathcal{P}_{c_i})\mid i=1,\ldots,m\}$, where each $c_i$ denotes a concept, $\mathcal{P}_{c_i}$ is the distribution of the images with the concept.
    \vspace{-5pt}
\end{definition}

We show examples in Figure~\ref{fig:demo} (Appendix~\ref{app:concept-bank}), where we utilize text-to-image generative models, \eg Stable Diffusion~\cite{RombachBLEO22} to generate concept images that represent $\mathcal{P}_{c_i}$, using the concept names as prompts. 

Moreover, the demand for interpretability calls into aligning concepts with the inner state of deep models. 
Without loss of generality, 
we denote a deep model as $f=h\circ g$, where $g$ is an encoder, and $h:\mathbb{R}^d\rightarrow \mathbb{R}^{|\mathcal{Y}|}$ is a linear layer.
Thus, given the model $f$ and a concept $c_i$, we define a query operation to extract the high-dimensional concept representation $v_i$. 
Concretely, we construct the positive set $I^{p}_{i}$ by sampling $N^p$ images from $\mathcal{P}_{c_i}$, and the negative set $I^{n}_{i}$ by sampling $N^n$ images randomly from $\{\mathcal{P}_{c_j}\mid j\neq i\}$. 
Following \citet{tcav}, we learn a linear SVM that finds a hyperplane in the hidden space $\mathbb{R}^d$ to best separate the positive images from the negative ones.
Then, we compute the vector $v_i$ orthogonal to the hyperplane as the Concept Activation Vector (CAV). 
Intuitively, a CAV is the direction in the hidden space representing the existence of a concept.
Thus, the concept bank serves as an auxiliary knowledge base to discover and mitigate the spurious correlation in the subsequent steps. 

\subsection{On Discovering Spurious Concepts}
\label{sec:search_shortcut}


With the concept bank, we aim to identify the spurious concepts from the concept candidates. 

\xhdr{Assumption on Training Distribution}
\rebuttal{
To guarantee the distinguishability of spurious concepts, we assume the training distribution is representative of the overall distribution. For example, if ``bed'' always coexists with ``cat'' in the training dataset, then there is no way we can distinguish that “cat” is the invariant concept while “bed” is not. Thus, the training distribution should reflect the essential characteristics as the overall distribution, which is implicitly assumed by the previous works~\cite{jtt,lff}. We formalize this assumption in Theorem~\ref{the:convergence}.}
\rebuttal{With the representativeness assumption, we further propose the following observation about an inconstancy property of spurious concepts:}

\vspace{2pt}
\begin{observation}[\textbf{Inconsistancy Property}]
    Spurious concepts tend to be present in heterogeneous subsets of the data and their correlations with the label are also heterogeneous.
\end{observation}
\vspace{-5pt}

For example, ``bed'' is not a common characteristic possessed by all the ``cat'' instances, it might be present in specific subsets of cat images and how much the presence of bed correlates with the cat label also differs across different subsets.
We want to leverage this property and insight to identify spurious concepts. 
Specifically, as a biased model correlates the label with spurious concepts, the distribution of spurious concepts has a large impact on the model's decision boundary. More importantly, this impact is often inconsistent across different environments. 
For example, imagine we cluster the ``cat'' instances based on whether it's an indoor or outdoor photo. 
The models trained under these two scenarios will exhibit distinct preferences for using the ``bed'' concept to make predictions.
Thus, the role of spurious concepts is highly fragile and sensitive in different environments, where the distribution of spurious concepts can change dramatically.
In contrast, \emph{invariant concepts}, \eg animal shape, are more uniform and homogeneous conditioned on any data environments.

Guided by this property, we propose a class-level metric, \textbf{Concept Sensitivity}, to indicate if a concept is spurious to a specific class. Here we introduce its computation steps:

\textbf{Step 1 (Clustering).}  We first seek good stratification on the training dataset to construct data environments. 
As explored by the previous works~\cite{george,bias-attribute}, even a biased model can well distinguish different features. 
\rebuttal{Thus, we leverage a model trained with ERM to generate both representations and predictions on the training instances, 
which, similar to \citet{Domino}, takes error type into account by including model predictions.}
Based the generated vectors, we conduct one-time clustering of the instances within each class. 
For each class $y\in\mathcal{Y}$, we obtain $G^{(y)}=\{G^{(y)}_j\}_{j=1}^{k}$, where $G^{(y)}_j$ is the $j$-th cluster in class $y$ and $k$ is the number of clusters. Finally, we construct the environments as 
\begin{equation}
\label{eq:envir}
G_j = {\textstyle\bigcup\limits_{y= 1}^{|\mathcal{Y}|}} G^{(y)}_j, \ \ j=1,\ldots,k,
\end{equation}
where the combination of clusters from each class is to avoid missing labels in the individual environments. 
Note that the combinations could be different according to the order of cluster indices, thus we randomize the cluster indices to update the environments for more robust training.


\textbf{Step 2 (Compute concept sensitivity).} 
For each $G_j$, we define the Environment Gradient Matrix (EGM) as

\begin{equation}
\label{eq:egm}
M_j =\nabla_{\omega}[\mathbb{E}_{(x, y)\sim G_j}\ell(f(x), y)]=\frac{\partial[\mathbb{E}_{(x, y)\sim G_j}\ell(f(x), y)]}{\partial\omega}
\end{equation}
where $\omega\in \mathbb{R}^{|\mathcal{Y}|\times d}$ is the parameter of $h$. 
Intuitively, $M_j$ represents the change in the parameter manifold given the observations in $G_j$ solely. While computing Equation~\ref{eq:egm} using all the training instances is expensive, we sample mini-batch data to approximate $M_j$. Further, to align the current model with the concept space, we query the concept bank to extract the concept representation $v_i$ of concept $c_i$. 
Then, we compute $v_i\cdot M_j^T \in \mathbb{R}^{|\mathcal{Y}|}$, which indicates how concept $c_i$ is preferred by each class under the environment $G_j$. 
For example, if {``bed''} is strongly correlated with {``cat''} in the environment $G_j$, then the updated decision boundary reflected by $M_j$ will align with the CAV of the {``bed''} concept with the logits output of being {``cat''}.
On top of this, we further define the concept sensitivity $S_i$ of concept $c_i$ as
\begin{align}
\label{eq:sens}
S_i &= \text{Var}(\{(v_i\cdot M_j^T)_{y_i'}\mid j=1,\ldots,k\}), \\
&\text{where}\ \ y_i'=\arg \max_{y}\ (\sum_{j=1}^k v_i\cdot M_j^T)_y\nonumber
\end{align}
\vspace{-18pt}

Here $y_i'$ is the class dominated by or strongly associated with the concept $c_i$. \text{Var} is the variance operater. 
For convenience, we refer to $(v_i\cdot M_j^T)_{y_i'}$ as Concept Tendency Score (CTS) of concept $c_i$ under the environment $G_j$. 
Thus, the concept sensitivity essentially evaluates the inconsistency of CTS in different environments.
A large variance of CTS indicates that the concept is unstable and its contribution to the final prediction varies across different environments. As the causal concept would exhibit invariance for environments \citep{irm, buhlmann2020invariance}, a large concept sensitivity can be interpreted as evidence of a concept being spurious and misleading in the model training. 
In our previous example where {``bed''} correlates with the {``cat''} class, the CTS's of {``bed''} in different environments show a large variance since {``bed''} has an inconstant existence with {``cat''} in the training data.
Also, note that the concept sensitivity is class-wise, this offers high-level interpretability to understand the spurious correlations in a certain class. 

\vspace{-5pt}
\subsection{Curing Bias via Concept-aware Intervention}
\label{sec:reduce_sens}
\vspace{-2pt}
Concretely, we blame the model bias for the imbalanced distribution of spurious concepts among classes.
For example, in Waterbirds~\cite{groupdro}, 95\% of instances in the \textit{landbird} class have \textit{land} backgrounds while only 4\% of instances in the \textit{waterbird} class involve with \textit{land}. 
Thus, such extreme imbalance encourages models to take advantage of spurious correlations as shortcuts to make predictions.

However, simply removing spurious attributes from the training dataset could introduce more noise and make the model overfit~\cite{KhaniL21}. 
Instead, we maintain the distribution balance of spurious concepts in different classes by data augmentation, to cancel the correlation.

\textbf{Step 3 (Concept-aware Intervention).} 
We denote each $H^{(y)}\in\mathbb{R}^{m}$ as a boolean vector where $H^{(y)}_i=1$ if $y_i'=y$, and $H^{(y)}_i=0$ otherwise.
Then we compute the concept probability on each class by normalizing the masked concept sensitivity, \ie $P^{(y)}=S\cdot H^{(y)}/\sum[S\cdot H^{(y)}]$.
Intuitively, the concept probability answers both ``what are the concepts correlated with the class $y$'' and ``how strong are their spurious correlations''.
To maintain the balance of spurious concepts, we sample concept images with probability $P^{(y)}$, and mix up them with the instances in their non-dominant classes. 
Formally, we have
\begin{align}
\label{eq:mixup}
    X' &= \lambda X + (1-\lambda) X_{(\mathcal{C}, P^{(y)})}, \ \ Y' = Y,
        \vspace{-15pt}
\end{align}
where $\lambda\sim {\rm Beta}(\beta_1, \beta_2)$. $X_{(\mathcal{C}, P^{(y)})}$ denotes concept images sampled with probability $P^{(y)}$ from concept bank $\mathcal{C}$. $(X, Y)$ are drawn from the subset where each $Y_i\in \mathcal{Y}/\{y\}$.
Intuitively, more sensitive concepts indicate a larger degree of imbalance between the dominant class and the other classes. 
And we devise such leave-one-out augmentation to compensate the imbalance, where concept images with high sensitivity are more frequently sampled to be mixed up, achieving the concept distribution balance.
In this way, the intervened dataset removes the spurious correlations involving multiple spurious concepts from the training dataset.

\begin{algorithm}[t]
    \small
    \caption{Pseudocode of \ourst}
    \label{alg:main}
    \begin{algorithmic}[1]
    \REQUIRE Training data $\mathcal{D}$, concept bank $\mathcal{C}$, a model $f=h\circ g$, learning rate $\alpha$, parameters $\beta_1, \beta_2$ of Beta distribution
    \STATE Obtain $\{\{G_j^{(y)}\}_{j=1}^k\}_{y=1}^{|\mathcal{Y}|}$ by clustering the image embeddings
    \WHILE{not converge} 
    \STATE Randomize cluster indices and obtain $\{G_j\}_{j=1}^k$ (Eq.~\ref{eq:envir})
    \STATE $\{P^{(y)}\}_{y=1}^{|\mathcal{Y}|}\leftarrow $\begin{small}$\textsc{Concept\_Sensitivity}(G, f, \mathcal{C})$\end{small}
    \FOR{each class $y$} 
    \STATE Sample minibatch $(X, Y)$, where each $Y_i\neq y$
    \STATE Sample concept images $X_{(\mathcal{C}, P^{(y)})}$ with prob. $P^{(y)}$
    \STATE Conduct mixup to obtain $(X', Y')$ (Eq.~\ref{eq:mixup})
    \STATE  $\phi \leftarrow \phi - \alpha\cdot\partial\ell[\mathbb{E}(f(X'), Y')]/\partial\phi$ 
    \ENDFOR
    \ENDWHILE
    \STATE 
    \STATE \textbf{function} \textsc{Concept\_Sensitivity}{($G, f, \mathcal{C}$)}
    \STATE Query for the CAV matrix $V=[v_1^T, \ldots, v_{m}^T]$ 
    \FOR{$j = 1,\ldots, k$}
    \STATE Sample minibatch $(X, Y) \sim G_j$
    \STATE Compute the EGM $M_j$ (Eq.~\ref{eq:egm})
    \ENDFOR
    \FOR{each concept $c_i$}
    \STATE Compute the dominant class $y_i'$ and sensitivity $S_i$ (Eq.~\ref{eq:sens})
    \STATE $H^{(y)}_i \leftarrow \mathbb{I}(y_i'=y)$, $\ \ y=1,\ldots, |\mathcal{Y}|$
    \ENDFOR
    \STATE \textbf{return} $\{P^{(y)}\}_{y=1}^{|\mathcal{Y}|}$
    \end{algorithmic}
\end{algorithm}

\xhdr{Adaptive Mitigation}
However, the model can learn various spurious correlations at different stages of training.
It is necessary for the concept sensitivity to be adjusted accordingly to thoroughly correct the model's decision boundary. 
Therefore, as shown in Figure~\ref{fig:framework}, we propose an adaptive framework \ourst, which iteratively conducts spurious concept discovery (Step 2) and concept-aware intervention (Step 3). 
At each epoch, \ours computes concept sensitivity which guides the concept-aware intervention. Then, the model evolves on the newly intervened dataset, where the spurious correlations are canceled. 
Thus, \ours can gradually mitigate the spurious correlations learnt by the model in the previous training.
In Theorem~\ref{the:convergence} of Section 5, we provide guarantees for the convergence of concept sensitivity and the final model. By iteratively mixing up images as in Equation~\ref{eq:mixup}, \ours reduces the contribution of spurious concepts to the final model and improves model generalization. 



\begin{table*}[t]
\small
\caption{Overall experimental results. The best results are \textbf{bold} and the second best results are \underline{underlined}. }
\label{tab:main}
\begin{center}
\begin{tabular}{l|cc|cc|cc|c}
\toprule
\multirow{2}{*}{} & \multicolumn{2}{c|}{MetaShift} & \multicolumn{2}{c|}{Waterbirds} & \multicolumn{2}{c|}{FMoW} & ISIC  \\
& Avg. Acc. & Worst Acc. & Avg. Acc. & Worst Acc.& Avg. Acc. & Worst Acc. & Avg. AUROC \\\midrule
ERM       & 72.9 $\pm$ 1.4\% & 62.1 $\pm$ 4.8\%       & 97.0 $\pm$ 0.2\% & 63.7 $\pm$ 1.9\%   & 53.0 $\pm$ 0.6\% & 32.3 $\pm$ 1.3\%         &   36.4 $\pm$ 0.7\% \\
ERM+aug &  75.5 $\pm$ 1.7\%     & 65.7 $\pm$ 3.3\%   & 87.4 $\pm$ 0.5\% &76.4 $\pm$ 2.0\%  & 55.5 $\pm$ 0.4\% &    \underline{35.7 $\pm$ 0.3\%}    & 38.9 $\pm$ 1.5\%\\
UW        & 72.1 $\pm$ 0.9\% &  60.5 $\pm$ 3.8\%      &  96.3 $\pm$ 0.3\%& 76.2 $\pm$ 1.4\%   &   52.5 $\pm$ 0.5\%   &  30.7 $\pm$ 1.5\%  &  39.2 $\pm$ 0.6\%\\
IRM& 73.9 $\pm$ 0.8\% & 64.7 $\pm$ 2.1\%       & 87.5 $\pm$ 0.7\% & 75.6 $\pm$ 3.1\%                & 50.8 $\pm$ 0.1\% & 30.0 $\pm$ 1.4\%           &  \underline{45.5 $\pm$ 3.6\%} \\
IB-IRM& 74.8 $\pm$ 0.2\%& 65.6 $\pm$ 1.1\%       & 88.5 $\pm$ 0.9\% & 76.5 $\pm$ 1.2\%                & 49.5 $\pm$ 0.5\% & 28.4 $\pm$ 0.9\%    &  38.6 $\pm$ 1.5\%  \\
V-REx&  72.7 $\pm$ 1.7\%  & 60.8 $\pm$ 5.5\%       & 88.0 $\pm$ 1.4\% & 73.6 $\pm$ 0.2\%& 48.0 $\pm$ 0.6\% & 27.2 $\pm$ 0.8\%&   24.5 $\pm$ 6.4\%\\
CORAL& 73.6 $\pm$ 0.4\% & 62.8 $\pm$ 2.7\%       & 90.3 $\pm$ 1.1\% & 79.8 $\pm$ 1.8\%                & 50.5 $\pm$ 0.4\% & 31.7 $\pm$ 1.2\%          &    37.9 $\pm$ 0.7\% \\
Fish& 64.4 $\pm$ 2.0\% & 53.2 $\pm$ 4.5\%       & 85.6 $\pm$ 0.4\% & 64.0 $\pm$ 0.3\%                & 51.8 $\pm$ 0.3\% & 34.6 $\pm$ 0.2\% &  42.0 $\pm$ 0.8\%\\
GroupDRO& 73.6 $\pm$ 2.1\% & \underline{66.0 $\pm$ 3.8\%}       & 91.8 $\pm$ 0.3\% &\textbf{90.6 $\pm$ 1.1\%}        & 52.1 $\pm$ 0.5\% & 30.8 $\pm$ 0.8\%&   36.4 $\pm$ 0.9\%\\
JTT& 74.4 $\pm$ 0.6\% & 64.6 $\pm$ 2.3\%& 93.3 $\pm$ 0.3\%& 86.7 $\pm$ 1.5\% & 52.5 $\pm$ 0.3\% &33.4 $\pm$ 0.9\% & 33.8 $\pm$ 0.0\% \\
DM-ADA& 74.0  $\pm$ 0.8\% & 65.7 $\pm$ 1.4\%       & 76.4 $\pm$ 0.3\% & 53.0 $\pm$ 1.3\%                & 51.6 $\pm$ 0.2\% & 34.2 $\pm$ 0.8\% &  35.8 $\pm$  1.0\%\\
LISA& 70.0 $\pm$ 0.7\% & 59.8 $\pm$ 2.3\%       & 91.8 $\pm$ 0.3\% & 88.5 $\pm$ 0.8\% & 52.8 $\pm$ 0.9\% & 35.5 $\pm$ 0.7\%  &  38.0 $\pm$ 1.3\% \\
\midrule
\textbf{\ourst} & 75.5 $\pm$ 1.1\%  & \textbf{73.5 $\pm$ 1.4\%} & 93.8 $\pm$ 0.7\% & \underline{88.7 $\pm$ 0.4\%} &  53.9 $\pm$ 0.4\%&  \textbf{36.1 $\pm$ 1.8\%}&  \textbf{55.1 $\pm$ 2.3\%}\\
\bottomrule
\end{tabular}
\end{center}
\vspace{-1.5em}
\end{table*}

\section{Experiments}
\label{sec:experiments}

We aim to answer the following research questions:
\vspace{-10pt}
\begin{itemize}[leftmargin=*]
    \item \textbf{RQ1}: How effective is  \ours  on tasks with spurious correlations, compared to state-of-the-art baselines?
    \vspace{-5pt}
    \item \textbf{RQ2}: What are the training dynamics and insights of \ours that are beneficial for future works?
    \vspace{-5pt}
    \item \textbf{RQ3}: How does each component affect \ourst's performance and contribute to its improvements?
\end{itemize}

\vspace{-10pt}
\subsection{Settings}

\vspace{-3pt}
\xhdr{Datasets} 
We summarize the datasets in Appendix~\ref{app:datasets}. 
We consider image classification tasks with various types of spurious correlations. 
Specifically, Waterbirds~\cite{groupdro} associates each class with water or land backgrounds, and MetaShift~\cite{metashift} constructs disjoint spurious attributes for each class. We also use FMoW from Wilds Benchmark~\cite{wilds} where satellite images are collected from different geographical regions that contribute to potential spurious correlations. 
Moreover, we consider a challenging task, ISIC~\cite{isic}, which classifies dermoscopic images of skin lesions into benign or melanoma. We use the train-test splits in \citet{isic-split}, where each training split amplifies the correlations with 7 spurious attributes. 
This task is difficult because multiple spurious attributes, \eg hairs and gel bubbles, could co-exist and cover the skin lesion region. 

\vspace{-2pt}
\xhdr{Baselines}
We compare \ours with Empirical Risk Minimization (ERM) with and without data augmentations; 
Upweighting (UW) which upweights the instances of minority groups; 
Invariant Learning algorithms: IRM~\cite{irm}, IB-IRM~\cite{ibirm}; Domain generalization/adaptation methods: V-REx~\cite{rex}, CORAL~\cite{coral}, and Fish~\citep{fish}; Instance reweighting methods: GroupDRO~\cite{groupdro}, JTT~\citep{jtt}; Data augmentation methods: DM-ADA~\cite{dm-ada}, LISA~\citep{lisa}. 
  
\vspace{-2pt}
\xhdr{Concept Bank} Inspired by previous works~\cite{cimpoi14, FongV18}, we build a concept bank with 224 concepts under 6 categories.
We generate 200 images per concept from a pre-trained text-to-image generation model, Stable Diffusion~\cite{RombachBLEO22}. 
To avoid unrealistic interventions,
we select the concept categories for each dataset as shown in Table~\ref{tab:category}. 
The details of concept bank construction and concept category selection are described in Appendix~\ref{app:concept-bank}. 

\vspace{-2pt}
\xhdr{Model Training} 
We summarize the hyperparameters in Appendix~\ref{app:models} and use Gaussian Mixture Model (GMM) as the clustering algorithm.
In Waterbirds, due to the extreme imbalance of majority and minority groups, we upweight the minority group for more stable results. While we do not require group information on the other datasets in training.

\vspace{-2pt}
 \xhdr{Evaluation} For ISIC, since the group size is $2^7$ considering combinations of spurious attributes, which results in many small groups, we compute the average AUROC score across the train-test splits, as standard in~\citet{isic-split}. 
 For other datasets, we evaluate the average and worst-group performance. 
 All the experiments are repeated three times.

\vspace{-5pt}
\subsection{Overall Results (RQ1)}

\xhdr{Analysis on the baselines}
In Table~\ref{tab:main}, we summarize the overall performance of \ours and the baselines. 
We observe that ERM with 
 data augmentations 
 constantly surpasses ERM, showing the effect of simple data augmentations in preventing the model from overfitting.
Also, we found that GroupDRO performs well in MetaShift and Waterbirds datasets. Yet, its performances are close to or worse than ERM in FMoW and ISIC datasets, which is in line with the observation in \citet{domainbed, wilds} that GroupDRO generally fails to improve over ERM in the wild.
Moreover, the models trained under invariant learning are suboptimal given the insufficient performance. 

Furthermore, we focus on the ISIC dataset. 
One interesting observation is that JTT fails -- the average AUROC is 3\% less than ERM -- which also justifies our assumption that the effectiveness of instance reweighting is largely limited when the minority instances are rare. 
Under this setting, the results of data mixup strategies are also unsatisfactory. 
Specifically, intra-label mixup proposed by LISA can cause even stronger spurious correlations, as it increases the population of majority groups
. 
In conclusion, the limitations of baselines prevent them from obtaining steady success.

\xhdr{The effectiveness of \ourst}
Overall, \ours outperforms most of the baselines in terms of the worst group accuracy. 
In particular, we obtain large performance gains over the best baselines in MetaShift and ISIC datasets by 7.5\% and 9.6\%, respectively. 
This evidence suggests that \ours is able to mitigate bias when combinations of spurious attributes exist. 
In FMoW, \ours achieves the state-of-the-art result on both average accuracy and worst-group accuracy, showing that our method can also perform well in wild image datasets.
In Waterbirds, \ours improves over JTT by 2.7\% while it underperforms GroupDRO. A potential explanation is that the images in the concept bank do not exactly cover the spurious attributes, which hinders the strength of mitigation. Nevertheless, \ours provides interpretability to understand model bias, which is discussed in Section~\ref{sec:interp}.

\subsection{Interpretability and Training Dynamics (RQ2)}
\label{sec:interp}

Besides the excellent task performance, our algorithm provides high transparency thanks to concept sensitivity.
Here we validate that concept sensitivity correctly reveals spurious correlation and enables user-oriented understanding of the spurious correlations during training.

\begin{figure}[t]
    \centering    
    \includegraphics[width=0.42\textwidth]{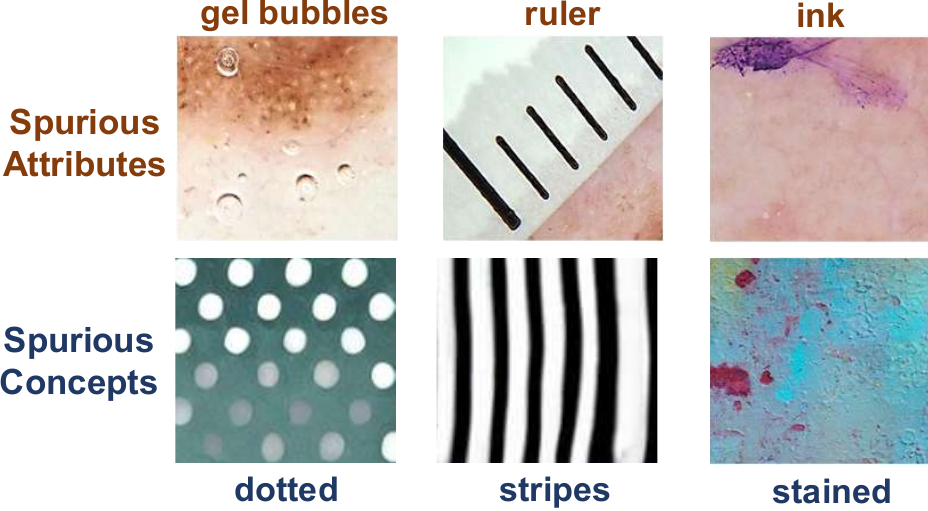}
  \vspace{-8pt}
  \caption{Alignment of spurious attributes and spurious concepts.}
  \label{fig:isic-compare}
\end{figure}

\xhdr{Validation on the  interpretations of DISC}
We investigate whether the concept sensitivity faithfully reflects the spurious correlations in the training data. For each concept, we compute the cumulative concept sensitivity over the training epochs to indicate the degree of overall spurious correlation.
%
The top 3 concepts with the largest cumulative sensitivity are \textbf{``dotted''} (0.056), \textbf{``stripes''} (0.032), and \textbf{``stained''} (0.030). Meanwhile, we borrow Cramér's V ~\citep{cramer2016mathematical} to measure the spuriousness for the 7 ground truth spurious attributes, where \textbf{gel bubbles} (0.184), 
\textbf{ruler} (0.411),
\textbf{ink} (0.215) are among the top spurious attributes. Interestingly, in Figure~\ref{fig:isic-compare}, we found strong alignments between the spurious concepts and the spurious attributes. For example, the rulers and ``stripes'' have a large feature-level similarity. 

The conclusion here is two-fold. First, the interpretations align well with the ground-truth spurious attributes, showing their trustworthiness. \rebuttal{We also provide a qualitative comparison of the interpretations of DISC and the existing interpretability methods in Appendix~\ref{app:interp} to further show the high quality of DISC interpretations.}
 Second, we demonstrate our method's applicability when certain concepts are absent from the concept bank, \eg ruler, which are substituted by other concepts preserving the same essential attributes. 
Such global and unambiguous interpretation clearly reveals the spurious correlations. 

\rebuttal{\xhdr{Robustness of interpretations and mitigation under absent ground truth spurious concepts} 
The previous example shows that DISC finds highly similar concepts even when the ground truth concepts are not included in the concept bank. 
Thus, we aim to investigate whether such a pattern is consistent. We designed two experiments: (1) Removing ``sofa'' and ``bed'' concepts (correlated with ``cat'') on MetaShift, and (2) removing ``bamboo'' and ``forest'' concepts (correlated with ``landbird'') on Waterbirds. We run the DISC algorithm under concept removal for each case and observe the interpretations on the corresponding class before and after the removal. }

\rebuttal{The top 3 interpretations before and after removal are \textbf{(wrinkle, bed, curtain) $\rightarrow$ (fireplace, bedrooms, paisley)} on MetaShift and \textbf{(bamboo, forests, flowerpot)$\rightarrow$(canopy, ground, plant)} on Waterbirds. Interestingly, we find that the interpretations before and after the removal have some conceptual overlappings (\eg \textbf{``bed''$\rightarrow$
''bedrooms''} on MetaShift, \textbf{``forests'' 
 $\rightarrow$``canopy''} and \textbf{``flowerpot''$\rightarrow$
 ``plant''} on Waterbirds). We further study the effect of concept removal on the worst group performance. Concretely, the performance decreases by 0.2\% on MetaShift and 0.9\% on Waterbirds. The absent concepts have a minor effect on MetaShift. While the performance on Waterbirds dataset is more sensitive to the absent concepts, the performance after removal still outperforms most of the baselines.
}

\rebuttal{
\ours can discover spurious correlations when the CAV of the absent spurious concept is similar to the CAVs of other concepts. The intuition is that the spurious concept (\eg forests) and other concepts (\eg canopy) may share part of the essential attributes (\eg leaves or greenness) that partially cause the spurious correlation, which results in the similar CAVs in the embedding space. 
Thus, the interpretations and performance of \ours are robust when ground truth is missing, as supported by our experiments.
}

\begin{figure}[t]
    \centering    
    \includegraphics[width=0.48\textwidth]{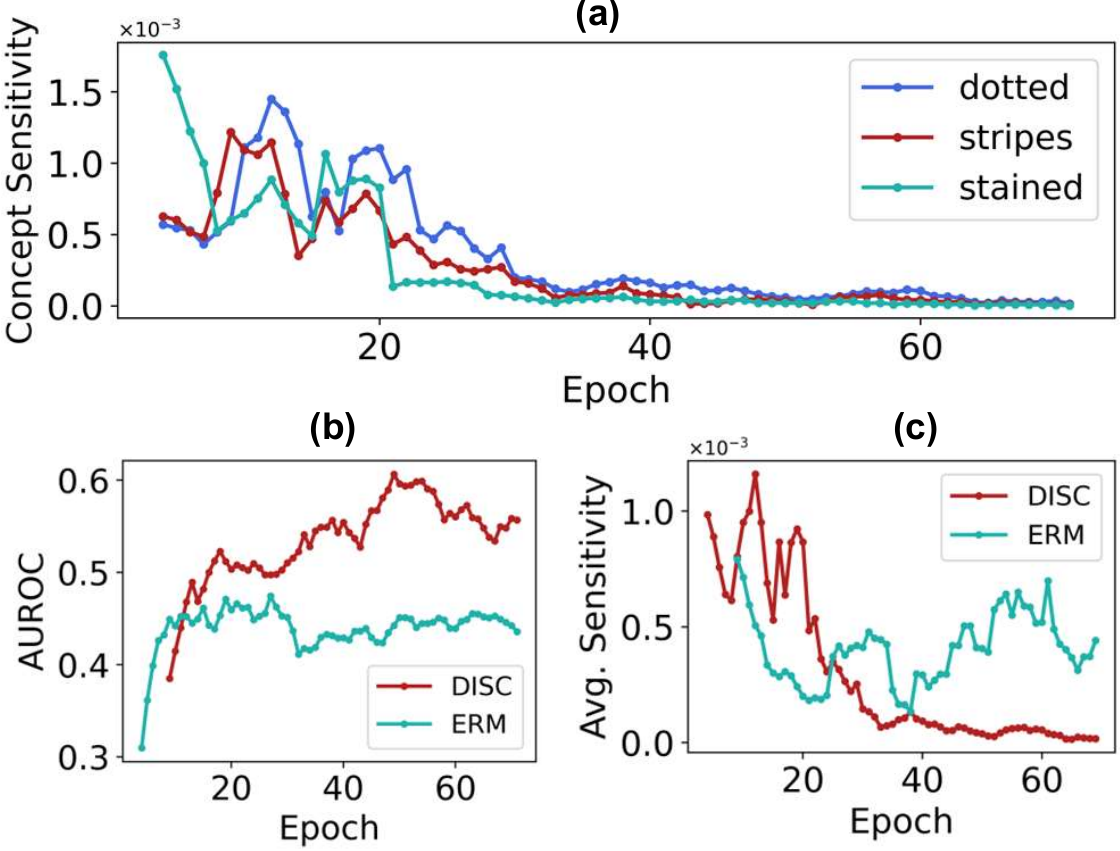}
    \vspace{-5pt}
    \caption{Training dynamics on ISIC. (a) Individual concept sensitivity \textit{vs.} epoch. (b) Test AUROC and (c) The average concept sensitive of \ours and ERM during training.}
    \label{fig:isic}
\end{figure}

\xhdr{Training dynamics as reducing concept sensitivity} 
The concept sensitivity reflects the extent of a model being affected by spurious bias, which helps probe the current model state.
As shown in Figure~\ref{fig:isic} (a), we observe the individual concept sensitivity during training. 
With randomly initialized weights, the model tends to learn from the spurious attributes at the early stage.
Correspondingly, the sensitivities of the top 3 concepts are relatively large at the beginning. 
Fortunately, we maintain the balance of spurious concepts among environments by concept-aware intervention, which gradually decreases the average sensitivity to almost zero. 

\begin{figure}
    \centering \includegraphics[width=0.485\textwidth]{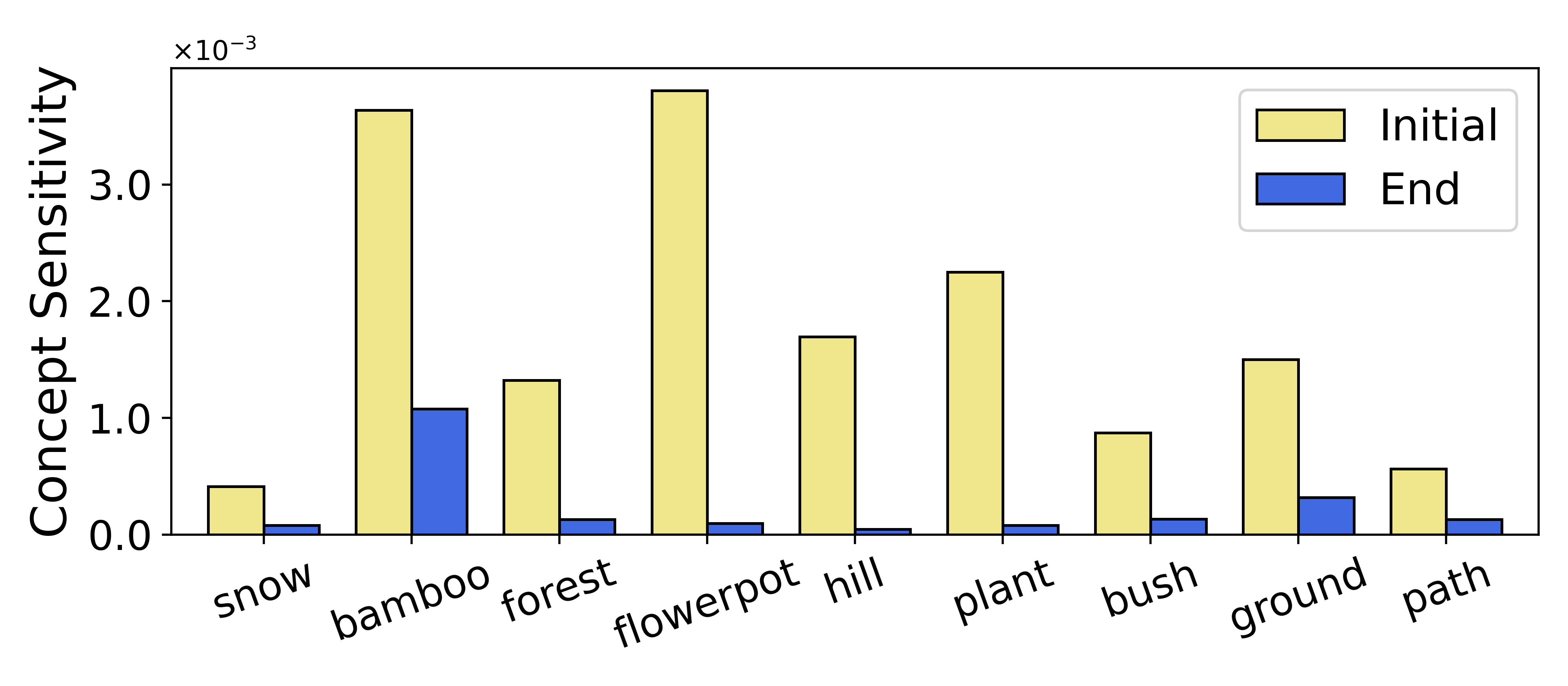}
    \vspace{-10pt}
    \caption{The concept sensitivity of spurious concepts on \textit{landbird} class on Waterbirds at the beginning and the end of training.}
    \label{fig:waterbirds}
\end{figure}

Moreover, we compare the average sensitivity of the three concepts and task performance for \ours and ERM. In Figure~\ref{fig:isic} (b) and (c), we found the average concept sensitivity of ERM has increased and remains high at the end. We believe the spurious concepts that are falsely associated with labels and remembered by the model result in the poor performance of ERM. 
In contrast, \ours reached a low concept sensitivity at the end. This pattern is consistent in the used datasets. For another example, in Figure~\ref{fig:waterbirds}, we show the comparison of the concept sensitivity before and after the training on Waterbirds.
The reduction of average sensitivity indicates that the model weight has reached a ``sweet spot'' where the model is not affected by spurious concepts. 



\subsection{Ablation and Sensitivity Study (RQ3)}
\label{sec:ablation_sens}
Here we empirically dissect the contribution of (1) concept sensitivity, 
(2) concept-aware Intervention , and \rebuttal{(3) adaptive mitigation in our algorithm}. 
We proposed three ablations models respectively: 
\begin{itemize}[leftmargin=*]
    \vspace{-4pt}
    \item \textbf{\ourst-Randint}, which discards  the concept sensitivity and randomly samples concept images for the cure step. 
    \vspace{-4pt}
    \item \textbf{\ourst-Reweight}, which replaces the cure step with reweighting instances unlikely to contain spurious concepts. Formally, the weight of an instance $(x_j, y_j)$ is 
    $\exp\{-\sum_{i=1}^m P^{(y_j)}_{i}\cdot\text{max}\{0, \frac{g(x_j)^T v_i}{|g(x_j)|\cdot|v_i|}\}\} $, which is negatively proportional to the alignment between its representation $g(x_j)$ and the CAVs of sensitive concepts.
    \vspace{-4pt}
    \item \rebuttal{\textbf{\ourst-Inadaptive}, which, instead of updating the concept sensitivity every epoch, generates the concept sensitivity based on the pre-trained ERM model and fixes it to conduct the intervention during training.}
\end{itemize}


\xhdr{Ablation results} In Table~\ref{tab:all-ablation} (Appendix~\ref{app:ablation_sens}), we report the ablation results. 
By comparing \ourst-Randint and \ourst, we discover that it is not just intervention, but the \textit{proper intervention} that can effectively reduce spurious correlation. 
By ``\textit{proper}'', we mean both ``what concept images should be chosen'' and ``what portion of training data should be intervened by a specific concept'', as fulfilled by concept sensitivity.
Further, the comparison 
    between \ourst-Reweight and \ours implies that the concept-aware intervention promotes the balance of spurious concepts and further mitigates spurious bias,  which is a key to \ourst's success. 
     \rebuttal{\ourst-Inadaptive consistently underperforms \ourst, and also underperforms DISC-Randint on FMoW and ISIC. Specifically, on these two datasets, we found while using fixed concept sensitivity scores removes the spurious correlations of the concepts with large sensitivity, the severity of the spurious correlations on the other concepts could increase, showing the importance of our adaptive mechanism in removing the spurious correlations more thoroughly.}
Overall, these three ablation models justify the efficacy of our framework. 

\begin{figure}[t]
    \centering
    \includegraphics[width=0.46\textwidth]{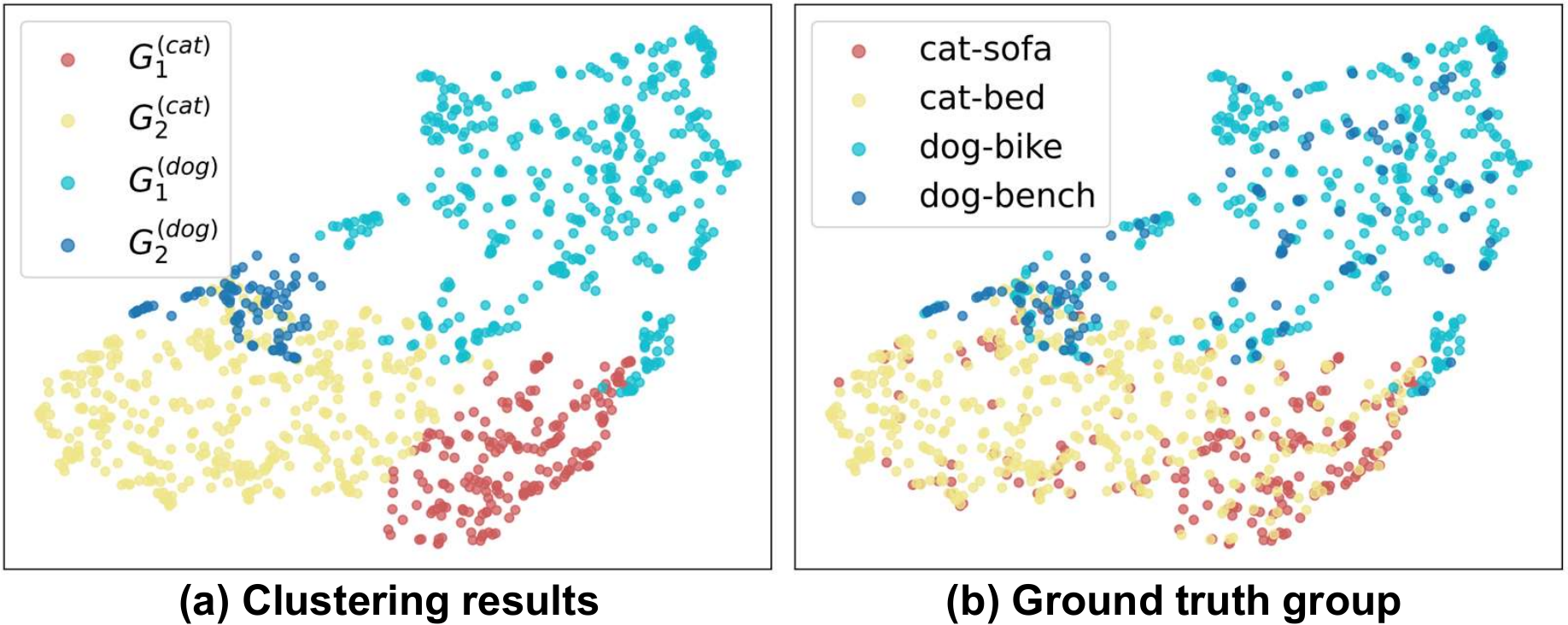}
    \vspace{-5pt}
    \caption{Visualization on the clustering and group assignments.
    }
    \label{fig:metashift}
\end{figure}

\xhdr{Unsupervised clustering} 
In Step 1, we conduct unsupervised clustering on the training instances to find good stratification on the spurious attributes. 
Here we are interested in \ourst's reliance on the clustering results. 
To search for the number of clusters $k$, we adopt Silhouette score as a heuristic following \citet{george}. 
Due to space limit, we include the results in Table~\ref{fig:all_sens} (Appendix~\ref{app:ablation_sens}). We show that, on most of the datasets, \ours outperforms the best baseline within a wide range of the cluster number. We further visualize the clustering results on MetaShift in Figure~\ref{fig:metashift}, where the clusters well match the ground truth group assignments. More visualizations are included in Appendix~\ref{app:ablation_sens}.
Thus, we validate the clustering by a trained ERM is informative to construct data environments.

\section{Theoretical Analysis}
\label{sec:theory}
In this section, we provide theoretical insights to understand the benefits of \ours in removing spurious correlation.

\xhdr{General Assumptions} We consider a Gaussian mixture model as the data-generating mechanism, which is widely adopted in the machine learning theory to shed light upon understanding complex phenomenon \citep{montanari2019generalization,liu2021self,ji2021power,deng2021adversarial,zhang2022and}. We consider the setting where the concepts are all well learned, and build model on the concept level.   
Specifically, the causal (invariant) concepts are modeled as 
    $x_{inv}=y\cdot\mu+\epsilon_{inv},$
where $y\in\{-1,1\}$ denotes the class index, $\mu\in\R^{p_1}$ represents the signal of the causal concept
with number of dimensions being $p_1$, and $\epsilon_{inv}$ is the noise term that is Gaussian with mean $\mathbf{0}$ and covariance matrix $\Sigma_1$. We further assume that the classes are balanced
, \ie $\Prob(Y=1)=\Prob(Y=-1)=1/2$. The distributions of the causal concepts are assumed to be invariant across training and test environments. 
In addition, the spurious concepts are modeled as 
$x_{spu}=\gamma_y^{(i)}+\epsilon_{spu}$, 
where $\gamma_y^{(i)}\in\R^{p_2}$ controls the spurious correlation and would vary according to different environments $i\in[k]$. As $x_{spu}$ is spurious, we have $\gamma=0$ in the test distribution $\mathcal P_{te}$. Similar to $\epsilon_{inv}$, $\epsilon_{spu}$ is the noise term and respects a Gaussian distribution with mean $\mathbf{0}$ and covariance matrix $I$. As each coordinate of $\mu$ and $\gamma$ represents a concept, we assume that  
the values of $\gamma_y^{(i)}$ are either 0 or 1, with 1 indicating the presence of the corresponding concept in the $i$-th environment of class $y$. 


 We consider minimizing the $\ell_2$ loss for the classification problem, which has been commonly used in the deep learning theory community \cite{ma2017diving,shankar2020neural,liang2021interpolating}.  
Here we compare the proposed \ours method with the standard ERM method 
\vspace{-5pt}
$$
(\hat\mu_{ERM},\hat\gamma_{ERM})=\arg\min_{\mu,\gamma}\sum_{i=1}^n (y_i-\mu^\top x_{inv,i}-\gamma^\top x_{spu,i})^2,
 $$
\vspace{-10pt}

 with the classifer being constructed as $\hat C_{ERM}(x)=sgn(\hat\mu_{ERM}^\top x_{inv}+\hat\gamma_{ERM}^\top x_{spu}).$
 Similarly, we define the classifier produced by \ours as $\hat C_{\ourst}(x)=sgn(\hat\mu_{\ourst}^\top x_{inv}+\hat\gamma_{\ourst}^\top x_{spu})$, where $\hat\mu_{\ourst}$ and $\hat\gamma_{\ourst}$ are the solution produced by Algorithm~\ref{alg:main}. 
 
 \begin{theorem}
 \label{the:convergence}
 Assuming that (1). $supp(\gamma_y^{(i)})$'s are disjoint for different $y$'s, and ${\rm Var}(\{\gamma_{y,j}^{(i)}\}_{i=1}^k)>K_0$ for $j\in[p_2]$ and some constant $K_0>0$, 
 (2).  $\|\mu\|_\infty\to 0$ when $p_1\to\infty$, and $K_1\le\lambda_{\min}(\Sigma_1)\le \lambda_{\max}(\Sigma_1)\le K_2$ for some constants $K_1,K_2>0$, (3). $p_1/n\to 0$ and $p_2$ is fixed. Then when training size $n$ is sufficiently large, Algorithm~\ref{alg:main} converges exponentially fast. 
 Moreover, with probability at least $1-o(1)$, the solution $(\hat\mu_{\ourst}, \hat\gamma_{\ourst})$ satisfies 
 $$
\Prob_{\mathcal P_{te}}[\hat C_{\ourst}(x)\neq y]<\Prob_{\mathcal P_{te}}[\hat C_{ERM}(x)\neq y].
$$
 \end{theorem}
 
\vspace{-5pt}
 We clarify the assumptions and include the detailed proof in Appendix~\ref{app:theory}. This theorem implies that by iteratively discovering and intervening, DISC mitigates the variation of the contribution of spurious concepts to the final model. Thus, \ours reduces the spurious correlations in the final model and outperforms ERM. 
\vspace{-5pt}
\section{Conclusion and Discussions}
\label{sec:con_fut}
\vspace{-3pt}
We propose \ours as a principled method to discover spurious correlations in a user-friendly way and then mitigate these correlations with data augmentation.  \ours is guided by the empirical observation that in many cases, spurious attributes are heterogeneous across different subsets of the data. Our systematic experiments demonstrate that \ours significantly improves model generalizability. Moreover, it provides useful insights into which concepts are sensitive and how this sensitivity changes over training. 

\rebuttal{
\xhdr{Limitations} While CAVs connect embedding space with concept space, the learning of the CAVs requires additional computation during training. Another limitation is that the concept bank using a generative model may have its own biases, which may limit the effectiveness of mitigation.}

\rebuttal{
\xhdr{Future Works} Interestingly, we conducted experiments on CIFAR-10-C and found DISC outperforms ERM by 13.1\% averaged across four types of corruptions, showing the potential of DISC in \textbf{OOD generalization}.
Moreover, future works can also build \textbf{better concept bank and tools for automatic concept category selection}, as discussed in Appendix~\ref{app:concept-bank}. One can also extend the \textbf{applicability of DISC} to multi-object vision datasets and NLP tasks or adopt DISC to analyze the concepts generated by techniques like SENN~\cite{senn}.
}

\section*{Acknowlegement}
\rebuttal{We thank Serina Chang, Zhi Huang, Lingjiao Chen, Boyang Deng, Ruocheng Wang, Yang Zheng at Stanford University who gave great suggestions to improve our work. 
We also thank the anonymous reviewers at ICML2023 conference for their insightful comments.}
The research of Linjun Zhang is partially supported by NSF DMS-2015378. The research of
James Zou is partially supported by funding from NSF CAREER and the Sloan Fellowship. 

\bibliography{ref}
\bibliographystyle{icml2023}
\onecolumn
\appendix

\section{Notation Table}
\label{app:notation}
\begin{table}[H]
\centering
\caption{Main notations used in the method section. Click \hyperref[sec:method]{\textcolor{teal}{here}} to return to the main paper.}
\vspace{2pt}
\label{tab:notation}
\begin{tabular}{cl}
\toprule
\textbf{Notation} & \textbf{Meaning}\\
\midrule
 $\mathcal{D}_{tr}/\mathcal{D}_{te}$ & Training/Testing dataset\\
 $\mathcal{P}_{tr}/\mathcal{P}_{te}$ & Training/Testing distribution\\
 $\mathcal{Y}$ & Label space\\
$f$&A deep model \\
 $g/h$&The encoder/last linear layer of $f$\\
 $\phi/\omega$ & Parameters of $f/h$\\
\midrule
 $k$& Number of clusters per class\\
 $m$& Number of concepts in the concept bank\\
 $d$ & Number of hidden dimensions\\
 \midrule
 $\mathcal{C}$ & A concept bank\\
 $c_i$ & The $i$-th concept in the concept bank $\mathcal{C}$\\
 $v_i$ & The concept activation vector of concept $c_i$\\
 $y'_i$ & The dominant class of concept $c_i$\\
 $\mathcal{P}_{c_i}$ & The distribution of the images from the $i$-th concept\\
 $S_i$ & The concept sensitivity of concept $c_i$\\
 $I^p_i/I^n_i$ & The positive/negative image set for concept $c_i$\\
 $N^p/N^n$ & The number of images in positive/negative image set\\
 \midrule
 $G_j^{(y)}$ & The $j$-th cluster in class $y$\\
 $G_j$ & The $j$-th environment\\
 $M_j$ & The Environment Gradient Matrix corresponding to $G_j$\\
 $H^{(y)}$ & A boolean mask of concepts for the dominant class $y$\\
 ${P}^{(y)}$ & Concept sampling distribution for class $y$\\
 $X_{(\mathcal{C}, {P}^{(y)})}$ & Images sampled from concept bank $\mathcal{C}$ with probability ${P}^{(y)}$\\
\bottomrule
\end{tabular}
\end{table}
\vspace{10pt}

\section{Theoretical Analysis}
\label{app:theory}


 \begin{theorem*}[\textbf{Restatement of Theorem~\ref{the:convergence}}]
  Assuming that (1). $supp(\gamma_y^{(i)})$'s are disjoint for different $y$'s, and ${\rm Var}(\{\gamma_{y,j}^{(i)}\}_{i=1}^k)>K_0$ for $j\in[p_2]$ and some constant $K_0>0$, 
 (2).  $\|\mu\|_\infty\to 0$ when $p_1\to\infty$, and $K_1\le\lambda_{\min}(\Sigma_1)\le \lambda_{\max}(\Sigma_1)\le K_2$ for some constants $K_1,K_2>0$, (3). $p_1/n\to 0$ and $p_2$ is fixed. Then when training size $n$ is sufficiently large, Algorithm~\ref{alg:main} converges exponentially fast. 
 Moreover, with probability at least $1-o(1)$, the solution $(\hat\mu_{\ourst}, \hat\gamma_{\ourst})$ satisfies 
\begin{equation*}
\begin{small}
\Prob_{\mathcal P_{te}}[\hat C_{\ourst}(x)\neq y]<\Prob_{\mathcal P_{te}}[\hat C_{ERM}(x)\neq y].
\end{small}
\end{equation*}
 \end{theorem*}

\xhdr{Clarification on the assumptions} We provide intuitive clarification on each of the assumptions as follows:
\begin{itemize}
    \item \textbf{(1)} The support operation $supp(\cdot)$ is a set consisting of all indices corresponding to nonzero entries in the input vector. The condition of the disjoint supports assumes that the spurious concepts are disjoint in different classes, which is supported by our observations in the experiments, \eg Figure~\ref{fig:metashift} in Section~\ref{sec:experiments}. The condition of the variance assumes that the strength of the spurious correlation (measured by the variance of the contribution of the spurious concepts) is not too small. This condition is necessary to detect spurious concepts by assuming a certain level of distinguishability.
    \item \textbf{(2)} $\lambda_{\text{min}}(\cdot)$ and $\lambda_{\text{max}}(\cdot)$ represent the smallest and largest eigenvalues of a matrix. The condition on $\|\mu\|_\infty$ assumes that the contribution of causal concepts should be spread out. The assumptions on the upper and lower bounds of eigenvalues of $\Sigma_1$ are standard in statistics and machine learning literature \cite{tony2019high,cai2021convex,cai2021cost,nakada2023understanding}.
    \item \textbf{(3)} We assume limited number of spurious concepts. Moreover, the number of training data $n$ needs to largely exceed the number of causal concepts so that the model can learn the invariant concepts well for the classification. 
\end{itemize}
 
 \begin{proof}
 We start the proof by analyzing the two main steps: concept sensitivity computation and the gradient update via mixup in each iteration. 

 Denoting the mini-batch size as $B$. 
 We note that we assume the concepts are all well-learned and only analyze the fitting on top of the well-learned concepts. The matrix multiplication by CAV will not show up throughout this proof. 
 Under our model set-up, at iteration $t$,  we have for $j\in[k]$, the $M_j$ in \eqref{eq:egm}
 equals to $$
M_{j,y}=X^\top(y \mathbf{1}_B-X(\mu_t;\gamma_t))/B.
$$

The corresponding parts for the causal and spurious features are respectively $$
M_{j,y;inv}=X_{inv}^\top(y \mathbf{1}_B-X(\mu_t;\gamma_t))/B=X_{inv}^\top y\textbf{1}_B/B-X_{inv}^\top X_{inv}\mu_t/B-X_{inv}^\top X_{spu}\gamma_t/B,
$$
and
$$
M_{j,y;spu}=X_{spu}^\top(y \textbf{1}_B-X(\mu_t;\gamma_t))/B=X_{spu}^\top y\textbf{1}_B/B-X_{spu}^\top X_{inv}\mu_t/B-X_{spu}^\top X_{spu}\gamma_t/B.
$$
As $X$ are assumed to be sub-gaussian, we have that \begin{align*}
M_{j,y;inv}=&\E[X_{inv}^\top y\textbf{1}_B/B-X_{inv}^\top X_{inv}\mu_t/B-X_{inv}^\top X_{spu}\gamma_t/B]+O(\sqrt\frac{p_1}{n})\\
=&\mu-(\mu\mu^\top+\Sigma_1)\mu_t-\gamma_t^\top\gamma_y^{(i)}\cdot\mu+O(\sqrt\frac{p_1}{n}),
\end{align*}
and 
\begin{align*}
M_{j,y;spu}=&\E[X_{spu}^\top y\textbf{1}_B/B-X_{spu}^\top X_{inv}\mu_t/B-X_{spu}^\top X_{spu}\gamma_t/B]+O(\sqrt\frac{p_2}{n})\\
=&y\cdot\gamma_y^{(i)}-\mu^\top\mu_t\cdot\gamma_y^{(i)}-(\gamma_y^{(i)}(\gamma_y^{(i)})^\top+I)\gamma_t+O(\sqrt\frac{p_2}{n})\\
=&y\cdot\gamma_y^{(i)}-\mu^\top\mu_t\cdot\gamma_y^{(i)}-(\gamma_y^{(i)})^\top\gamma_t\cdot \gamma_y^{(i)}+I\gamma_t+O(\sqrt\frac{p_2}{n}).
\end{align*}

Then, as we now consider the binary classification setting, the $S_i$ in \eqref{eq:sens} now equals to $$
S_{i,j}^{(y)}={\rm Var}(\{\gamma_{y,j}^{(i)}\}_{i=1}^k).
$$
Now, for the invariant part, as $\|\mu\|_\infty=O(1)$, and fixed $p_2$ implying that $|\gamma_t^\top \gamma_y^{(i)}|=O(1)$, we have for all $j\in[p_1]$, $$
S_{i,j}^{(y)}=o(1).
$$
Also, by assumption, we have $\|\gamma_y^{(i)}\|>1$ and ${\rm Var}(\{\gamma_{y,j}^{(i)}\}_{i=1}^k)>K_0$ for $j\in[p_2]$, and therefore for all $j\in[p_2]$ and some constant $K_0>0$, we have
$$
S_{i,j}^{(y)}=\Omega(1).
$$
As a result, with probability at least $1-o(1)$, the sampling according to $P^{(y)}$ will always draw from the spurious concepts from $\cup_{i=1}^k {\rm supp}(\gamma_y^{(i)})$. We denote such an event by $E$ with $\Prob(E)\ge 1-o(1)$.

Now we analyze the mixup part on the event $E$. According to our model setup, for $j\in[p_2]$, the concept image is modeled as the basis vector $e_j$, with the $j$-th entry equal to 1, indicating the presence of this concept. 

Letting $\tilde\gamma_y=\frac{1}{k}\sum_{i=1}^k\gamma_y^{(i)}$. Then after mixup, for the spurious concepts, there exists a vector $c_{y,spu}$ with support belonging to $\cup_{i=1}^k {\rm supp}(\gamma_y^{(i)})$ and nonzero entries are in $(0,1)$, such that the gradient update becomes $$
S_{spu}^{(i)}=\sum_{y\in\{-1,1\}}y\cdot(\tilde\gamma_y+c_{-y,spu})-\mu^\top\mu_t\cdot(\sum_{y\in\{-1,1\}}\tilde\gamma_y+c_{-y,spu})-(\sum_{y\in\{-1,1\}}(\tilde\gamma_y\tilde\gamma_y^\top+c_{-y,spu}c_{-y,spu}^\top)+I)\gamma_t+O(\sqrt\frac{p_2}{n}),
$$

Note that we assume $y\in\{-1,1\}$. Using the fact that the supports of $\tilde\gamma_y$ and $c_{-y,spu}$ are disjoint, we have that 
$$
S^t_{spu}=\sum_{y\in\{-1,1\}}y\cdot(\tilde\gamma_y+c_{-y,spu})-\mu^\top\mu_t\cdot(\sum_{y\in\{-1,1\}}\tilde\gamma_y+c_{-y,spu})-(\sum_{y\in\{-1,1\}}(\tilde\gamma_y+c_{-y,spu})(\tilde\gamma_y+c_{-y,spu})^\top+I)\gamma_t+O(\sqrt\frac{p_2}{n}).
$$
In addition, the gradient on the invariant (causal) part
$$
S^t_{inv}=\mu-(\mu\mu^\top+\Sigma_1)\mu_t-\gamma_t^\top\gamma_y^{(i)}\cdot\mu+O(\sqrt\frac{p_1}{n}).
$$
As a result, the update in each iteration $t$ of  is equivalent to running gradient descent on minimizing the loss function
$\ell(\hat\mu,\hat\gamma)=(\hat\mu;\hat\gamma)^\top \begin{pmatrix}\Sigma_1+\mu\mu^\top& 0 \\ 0&\sum_{y\in\{-1,1\}}(\tilde\gamma_y+c_{-y,spu})(\tilde\gamma_y+c_{-y,spu})^\top+I\end{pmatrix}(\hat\mu;\hat\gamma)+(\hat\mu;\hat\gamma)^\top (\mu;\sum_{y\in\{-1,1\}}y\cdot(\tilde\gamma_y+c_{-y,spu}))$.

Since $\lambda_{\min}(\Sigma_1), \lambda_{\min}(I)>K_1$, $\ell$ is a strongly convex function, implying that Algorithm~\ref{alg:main} converges exponentially fast. 

At last, we compare the performance of \ours and ERM.

Since $(\hat\mu_{\ourst},\hat\gamma_{\ourst})$ minimizes $\ell$, we can write out its analytical solution as $$
\hat\mu_{DISC}=(\Sigma_1+\mu\mu^\top)\mu+O(\sqrt\frac{p_1}{n}),
$$
and \begin{align*}
\hat\gamma_{DISC}=&(\sum_{y\in\{-1,1\}}(\tilde\gamma_y+c_{-y,spu})(\tilde\gamma_y+c_{-y,spu})^\top+I)^{-1}\sum_{y\in\{-1,1\}}y\cdot(\tilde\gamma_y+c_{-y,spu})\\
=&(\sum_{y\in\{-1,1\}}(\tilde\gamma_y)(\tilde\gamma_y)^\top+\sum_{y\in\{-1,1\}}(c_{-y,spu})(c_{-y,spu})^\top+I)^{-1}\sum_{y\in\{-1,1\}}y\cdot(\tilde\gamma_y+c_{-y,spu})
\end{align*}

Similarly, we have 
 $$
\hat\mu_{ERM}=(\Sigma_1+\mu\mu^\top)\mu+O(\sqrt\frac{p_1}{n}),
$$
and $$
\hat\gamma_{ERM}=(\sum_{y\in\{-1,1\}}(\tilde\gamma_y)(\tilde\gamma_y)^\top+I)^{-1}\sum_{y\in\{-1,1\}}y\cdot\tilde\gamma_y.
$$
Since all the entries of $\tilde\gamma_y$ are either 0 or 1, all the entries of $c_{y,spu}$ are between 0 and 1, and the support of $\tilde\gamma_y$ and $c_{-y,spu}$ are disjoint, we have that 
$$
\|\hat\gamma_{DISC}\|<\|\hat\gamma_{ERM}\|.
$$
Now we analyze the misclassification error in the test domain. For any $\hat\mu$ and $\hat\gamma$, we have \begin{align*}
\Prob_{\mathcal D_{te}}(sgn(\hat\mu^\top x_1+\hat\gamma^\top x_2)\neq y)=&\frac{1}{2}\Prob_{\mathcal D_{te}}(\hat\mu^\top\mu+\hat\mu^\top\epsilon_1+\hat\gamma^\top\epsilon_2>0)+\frac{1}{2}\Prob_{\mathcal D_{te}}(-\hat\mu^\top\mu+\hat\mu^\top\epsilon_1+\hat\gamma^\top\epsilon_2<0)\\
=&\frac{1}{2}\E[\Prob(\hat\mu^\top\mu+\hat\mu^\top\epsilon_1+\hat\gamma^\top\epsilon_2>0\mid \epsilon_1)]+\frac{1}{2}\E[\Prob(-\hat\mu^\top\mu+\hat\mu^\top\epsilon_1+\hat\gamma^\top\epsilon_2<0\mid \epsilon_1)]\\
=&\E[\Phi(-\frac{\hat\mu^\top\mu}{\sqrt{\|\hat\mu\|^2+\|\hat\gamma\|^2}})],
\end{align*}
where $\Phi$ is the cumulative distribution function of a standard normal distribution. As a result, using $\hat\mu^\top\mu=c\|\mu\|^2+O(\sqrt{p_1/n})>0$ and $
\|\hat\gamma_{DISC}\|<\|\hat\gamma_{ERM}\|,
$ We have that $$
\Prob_{\mathcal D_{te}}[\hat C_{\ourst}(x)\neq y]<\Prob_{\mathcal D_{te}}[\hat C_{ERM}(x)\neq y].
$$
 \end{proof}

\section{Datasets}
\label{app:datasets}
\begin{table}[H]
\caption*{\textit{(a)} \textbf{Metashift Dataset}.}
\begin{tabular}{cc llll|ll}
\toprule
\multirow{2}{*}{} &
\multicolumn{7}{r}{
\textbf{Target}: classify cat / dog.} \\
& \multicolumn{7}{r}{
\textbf{Spurious feature}: object / background; sofa, bed (cat); bench, bike (dog).} \\
\hline\\[-3mm]
\textbf{Image:} && 
\begin{tabular}{c}\includegraphics[width=0.075\textwidth]{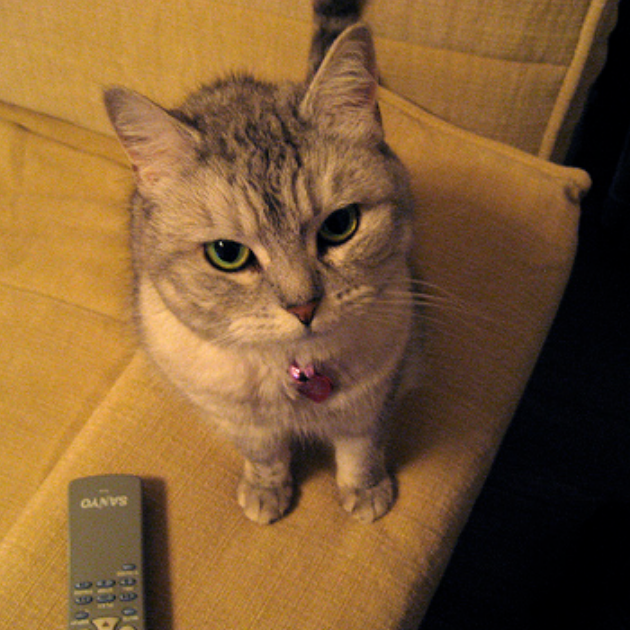}\end{tabular} &
\begin{tabular}{c}\includegraphics[width=0.075\textwidth]{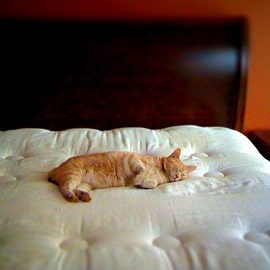}\end{tabular} &
\begin{tabular}{c}\includegraphics[width=0.075\textwidth]{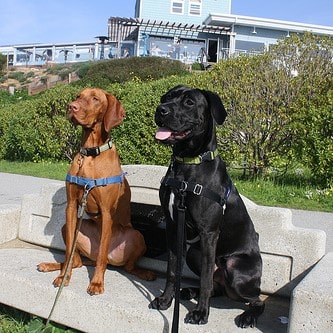}\end{tabular} &
\begin{tabular}{c}\includegraphics[width=0.075\textwidth]{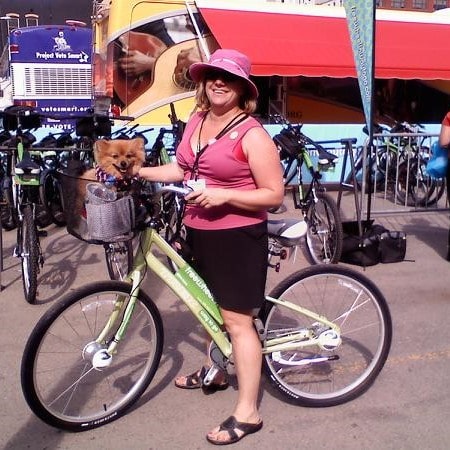}\end{tabular}&
\begin{tabular}{c}\includegraphics[width=0.075\textwidth]{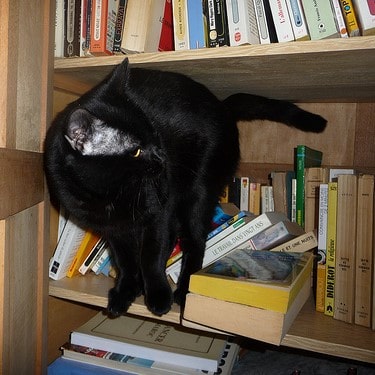}\end{tabular} &
\begin{tabular}{c}\includegraphics[width=0.075\textwidth]{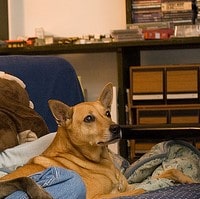}\end{tabular}\\
\rowcolor{mycham}
\textbf{Group $g$:} &\quad\quad& 
$0$
& $1$
& $2$
& $3$ 
& $4$
& $5$ 
\\
\textbf{Target $y\in\{0,1\}$:}    &&  0 (cat)& 0 (cat) & 1 (dog) & 1 (dog) & 0 (cat) & 1 (dog)\\
\rowcolor{mycham}
\textbf{Spurious $s$:}    && 0 (sofa) & 1 (bed) & 2 (bench) & 3 (bike) & 4 (shelf) & 4 (shelf)\\
\textbf{\# Train data:}  &&
 231&380&145&367&-&-\\
\rowcolor{mycham}
\textbf{\# Val data (OOD):}  
&& -&-&-&-&34&47\\
\textbf{\# Test data:}   
&& -&-&-&-&201&259\\
\bottomrule
\end{tabular}
\end{table}

\begin{table}[H]
\caption*{\textit{(b)} \textbf{Waterbirds Dataset.}}
\begin{tabular}{cc llll}
\toprule
 &
\multicolumn{5}{r}{
\textbf{Target}: bird type;\quad \textbf{Spurious feature}: background type.} \\
\hline\\[-3mm]
\textbf{Image:} && 
\begin{tabular}{c}\includegraphics[width=0.14\textwidth]{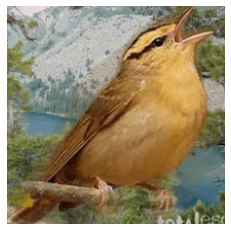}\end{tabular} &
\begin{tabular}{c}\includegraphics[width=0.14\textwidth]{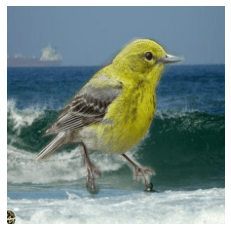}\end{tabular} &
\begin{tabular}{c}\includegraphics[width=0.14\textwidth]{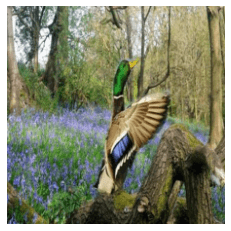}\end{tabular} &
\begin{tabular}{c}\includegraphics[width=0.14\textwidth]{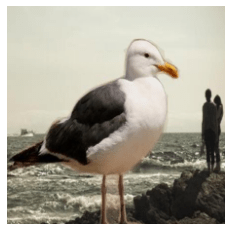}\end{tabular}
\\
\rowcolor{myblue}
\textbf{Group $g$:} &\quad\quad& 
$0$
& $1$
& $2$
& $3$ 
\\
\textbf{Target $y\in\{0,1\}$:}    && 0 (landbird) & 0 (landbird) & 1 (waterbird) & 1 (waterbird)
\\
\rowcolor{myblue}
\textbf{Spurious $s$:}    && 0 (land) & 1 (water) & 0 (land) & 1 (water)
\\
\textbf{\# Train data:}  &&
3,498 (73\%) &  184 (4\%) & 56 (1\%) & 1,057 (22\%)\\
\rowcolor{myblue}
\textbf{\# Val data:}    &&
467 & 466 & 133 & 133\\
\textbf{\# Test data:}    &&
2,255 & 2,255 & 642 & 642\\
\bottomrule
\end{tabular}

\end{table}


\begin{table}[H]
\caption*{\textit{(c)} \textbf{FMoW Dataset.}
}
\begin{tabular}{cc lllll}
\toprule
\multirow{2}{*}{} &
\multicolumn{6}{r}{
\textbf{Target}: one of 62 building or land use categories, e.g., park, shopping mall, dam, stadium, airport.} \\
& \multicolumn{6}{r}{
\textbf{Spurious features}: Unknown (not explicitly given by the data source). }\\
\hline\\[-3mm]
\textbf{Image:} && 
\begin{tabular}{c}\includegraphics[width=0.11\textwidth]{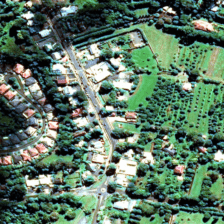}\end{tabular} &
\begin{tabular}{c}\includegraphics[width=0.11\textwidth]{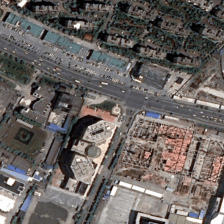}\end{tabular} 
& 
\begin{tabular}{c}\includegraphics[width=0.11\textwidth]{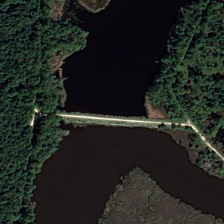}\end{tabular} &
\begin{tabular}{c}\includegraphics[width=0.11\textwidth]{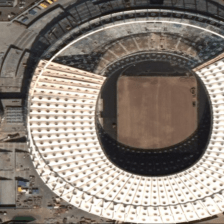}\end{tabular}&
\begin{tabular}{c}\includegraphics[width=0.11\textwidth]{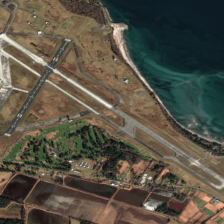}\end{tabular}\\
\rowcolor{mygreen}
\textbf{Group $g$:}    && Europe &  Asia &Americas  & Africa&Oceania\\
\textbf{\# Train data:}  &&
34,816&17,809&20,973&1,582&1,641\\
\rowcolor{mygreen}
\textbf{\# Val data:}    
&&7,732&4,121&6,562&803&693\\
\textbf{\# Test data:}  
&&5,858&4,963&8,024&2,593&666\\
\bottomrule
\end{tabular}
\end{table}

\begin{table}[H]
\caption*{\textit{(d)} \textbf{ISIC Dataset}. For methods that require domain information, we use the existence of hairs as the domain labels. 
Each training split amplifies different correlations, and the corresponding testing set provides reversed correlations.}
\begin{tabular}{cc ccccc}
\toprule
\multirow{2}{*}{} &
\multicolumn{6}{r}{
\textbf{Target}: benign / melanoma skin lesions} \\
& \multicolumn{6}{r}{
\textbf{Spurious features}: dark corners, hair, gel borders, gel bubbles, ruler, ink markings/staining, patches.}\\
\hline\\[-3mm]
\textbf{Image:} && 
\begin{tabular}{c}\includegraphics[width=0.13\textwidth]{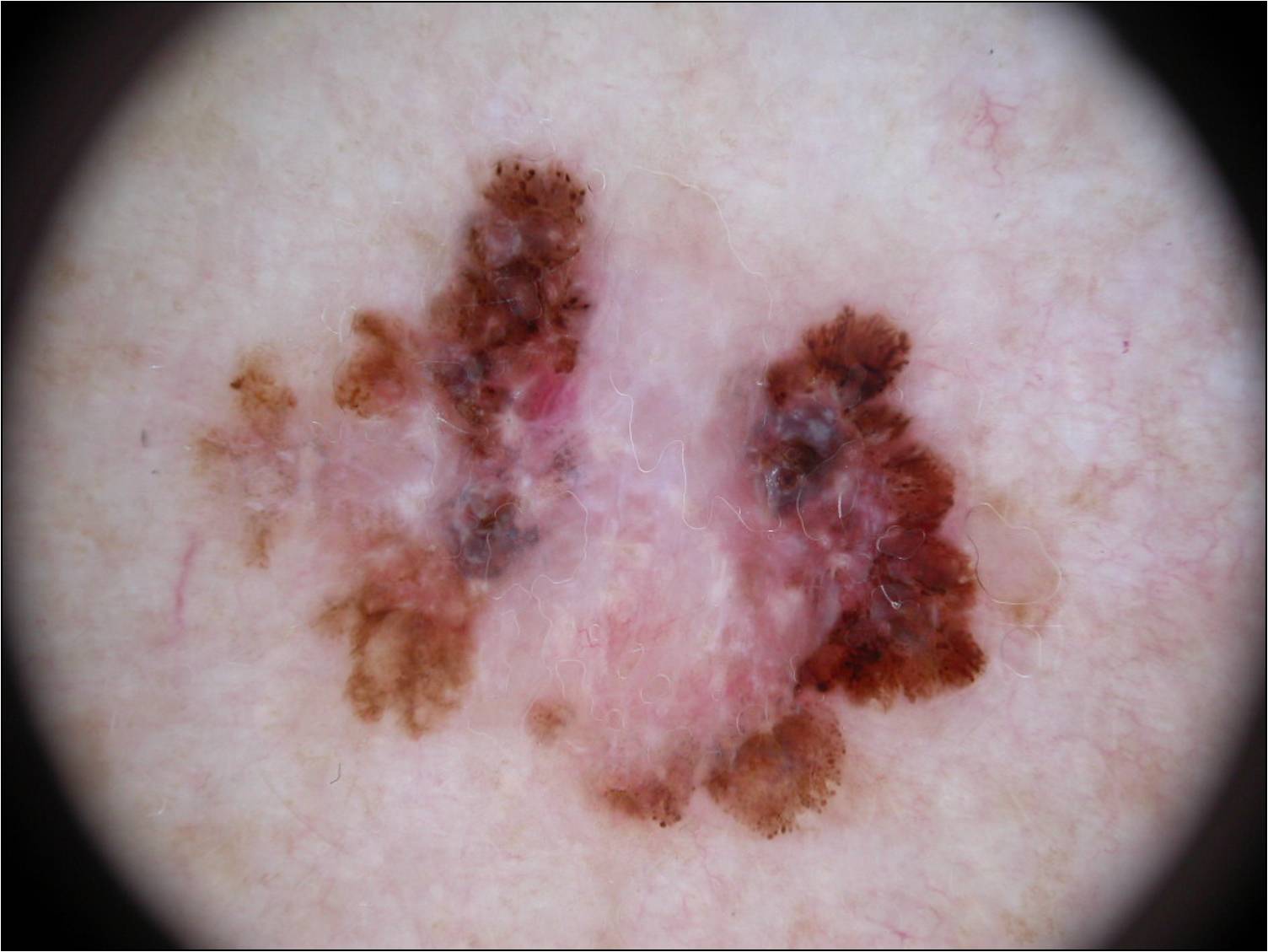}\end{tabular} &
\begin{tabular}{c}\includegraphics[width=0.13\textwidth]{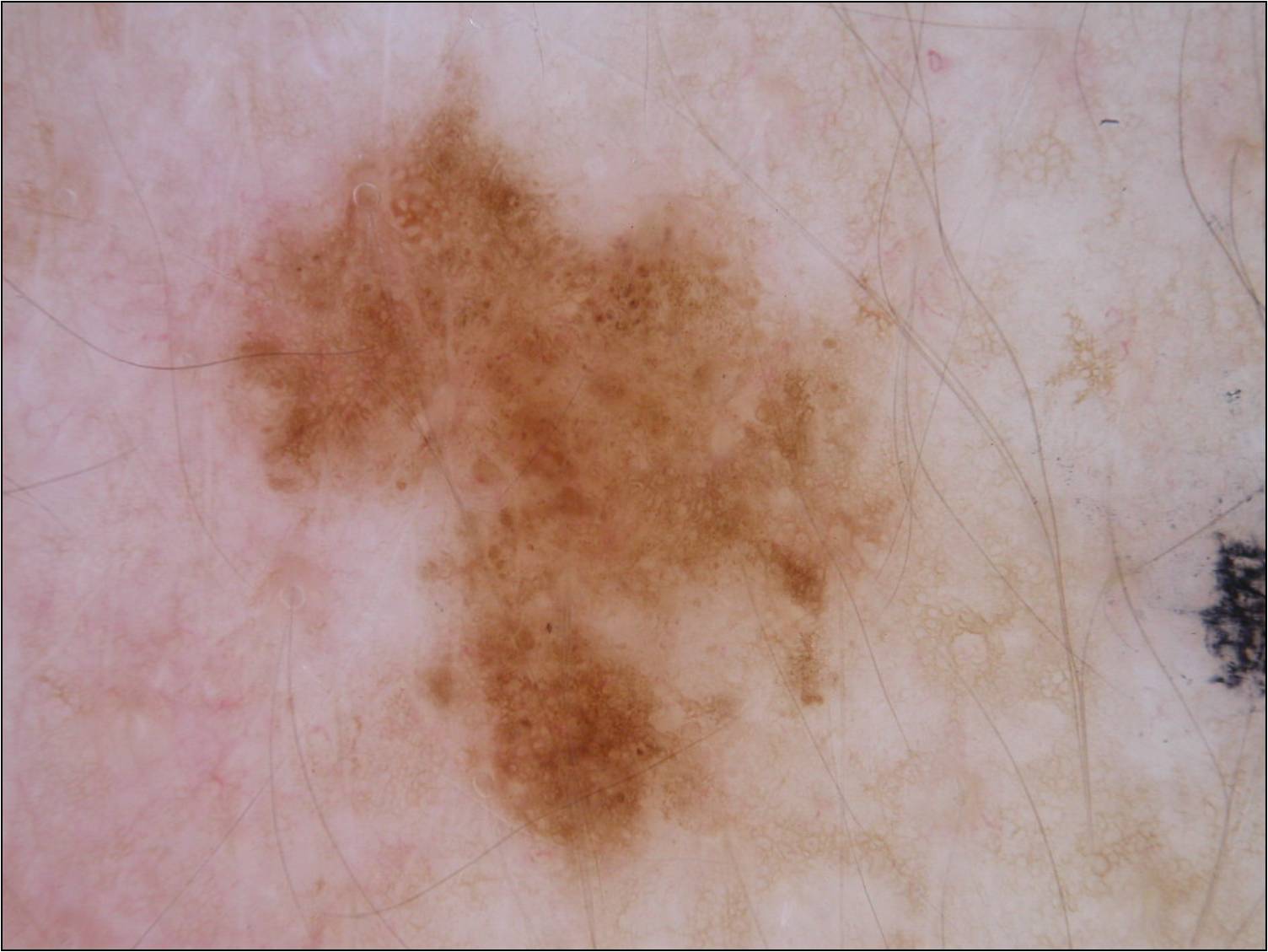}\end{tabular} &
$\ldots$&
\begin{tabular}{c}\includegraphics[width=0.13\textwidth]{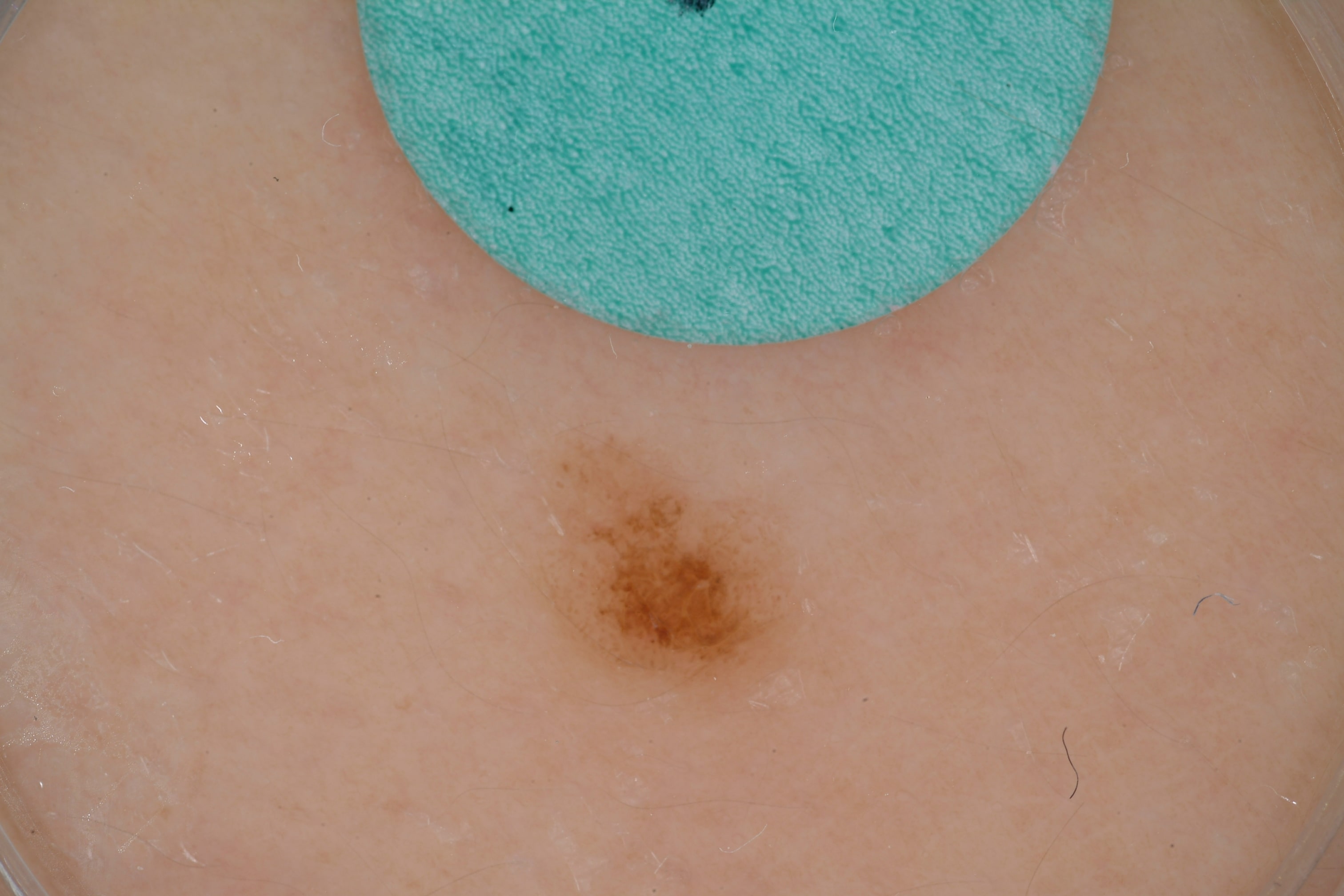}\end{tabular} &
\begin{tabular}{c}\includegraphics[width=0.13\textwidth]{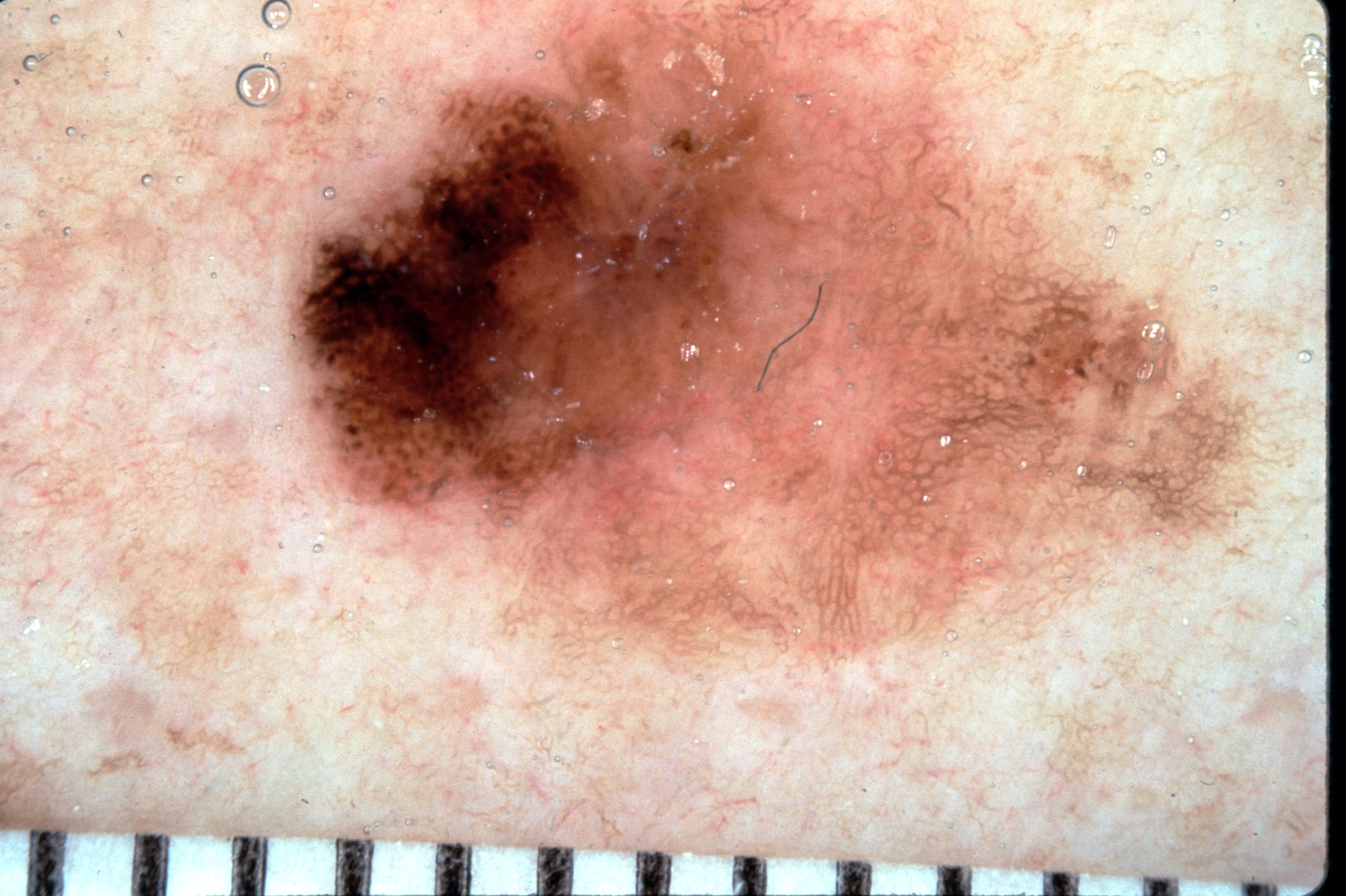}\end{tabular}\\
\rowcolor{mygray}
\textbf{Target $y\in\{0,1\}$:}    &&  0 (benign)& 0 (benign) &$\ldots$& 1 (malignant) & 1 (malignant)\\
\multirow{2}{*}{\textbf{Spurious $s$:}}     &&patch,  & ink,  &\multirow{2}{*}{$\ldots$}& dark corner,  & ruler, \\
&&gel border&hair&&gel bubble&dark corner\\
\rowcolor{mygray}
\textbf{\# Train data:}  &&
 \multicolumn{5}{c}{1,826}\\
\textbf{\# Val data:}    &&
 \multicolumn{5}{c}{154}\\
\rowcolor{mygray}
\textbf{\# Test data:} &&
\multicolumn{5}{c}{618 }  \\
\bottomrule
\end{tabular}
\end{table}

%
%

\section{Concept Bank}
\label{app:concept-bank}
\xhdr{Concept categories} In Table~\ref{tab:conceptbank}, we list all the 224 concepts in the concept bank under 6 categories, which are  \textit{(Color, Texture,
Nature, City, Household, Others)}. Note that the concept bank could be easily extended with user-defined concepts since the concept images are cheap to obtain, leveraging the text-to-image generative models. 

\begin{table}[H]
\centering
\caption{A comprehensive concept list of the concept bank in this work.}
\label{tab:conceptbank}
\begin{tabularx}{\textwidth}{lX}
\toprule
\textbf{Concept category} & \textbf{Concepts}\\
\midrule
\textbf{Color} &\small [\textit{blackness,
         blueness,
         greenness,
         redness,
         whiteness}]\\
\midrule
\textbf{Texture} &\small [\textit{concrete,
        granite,
        leather,
        laminate,
        metal,
        blotchy,
        blurriness,
        stripes,
        polka dots,
        knitted, cracked, frilly, waffled, scaly, lacelike, grooved, stratified, gauzy, marbled, 
        flecked, stained, braided, matted, meshed, cobwebbed, spiralled, dotted, crosshatched, wrinkled, 
        woven, potholed, crystalline, paisley, veined, fibrous, studded, bubbly, pleated, grid, 
        perforated, porous, interlaced, smeared, honeycombed, sprinkled, chequered, lined, banded, bumpy, 
        zigzagged, swirly, pitted, freckled}]\\
\midrule
\textbf{Nature} & \small[\textit{bamboo, 
        beach, 
        bridge, 
        bush, 
        canopy, 
        earth, 
        field,
        flower,
        flowerpot,
        fluorescent,
        forest,
        grass,
        ground,
        harbor,
        hill,
        lake,
        mountain,
        muzzle,
        palm, 
        path,
        plant,
        river, 
        sand,
        sea,
        snow,
        tree,
        water}]\\
\midrule
\textbf{City} &\small [\textit{awning, 
        base, 
        bench, 
        building, 
        earth, 
        fence, 
        field,
        ground,
        house,
        manhole, 
        path,
        snow,
        streets}]\\
\midrule
\textbf{Household} & \small[\textit{air-conditioner,
        apron,
        armchair,
        back-pillow,
        balcony,
        bannister,
        bathrooms,
        bathtub,
        bed,
        bedclothes,
        bedrooms,
        cabinet,
        carpet,
        ceiling,
        chair,
        chandelier,
        chest-of-drawers,
        countertop,
        curtain,
        cushion,
        desk,
        dining-rooms,
        door,
        door-frame,
        double-door,
        drawer,
        drinking-glass,
        exhaust-hood,
        figurine,
        fireplace,
        floor,
        flower,
        flowerpot,
        fluorescent,
        ground,
        handle,
        handle-bar,
        headboard,
        headlight,
        house,
        jar,
        lamp, 
        light, 
        microwave, 
        mirror, 
        ottoman,
        oven,
        pillow,
        plate,
        refrigerator,
        sofa,
        stairs,
        toilet}]\\
\midrule
\textbf{Others} &\small [\textit{bird,
        cat,
        cow,
        dog,
        horse,
        mouse,
        paw,
        arm,
        back,
        body,
        ear,
        eye,
        eyebrow,
        female-face,
        leg,
        male-face,
        foot,
        hair,
        hand,
        head,
        inside-arm,
        knob,
        mouth,
        neck,
        nose,
        outside-arm,
        ashcan,
        airplane,
        bag,
        bus, 
        beak,
        bicycle,
        blind,
        board,
        book,
        bookcase,
        bottle,
        bowl,
        box,
        brick,
        basket,
        bucket,
        bumper,
        can,
        candlestick,
        cap,
        car,
        cardboard,
        ceramic,
        chain-wheel,
        chimney,
        clock,
        coach,
        coffee-table,
        column,
        computer,
        counter,
        cup,
        desk,
        engine,
        fabric,
        fan,
        faucet,
        flag,
        floor,
        food,
        foot-board,
        frame,
        glass,
        keyboard,
        lid, 
        loudspeaker, 
        minibike, 
        motorbike,
        napkin,
        pack,
        painted,
        painting,
        pane,
        paper,
        pedestal,
        person,
        pillar,
        pipe}]\\
\bottomrule
\end{tabularx}
\end{table}

\begin{figure}[H]
    \centering
    \includegraphics[width=0.8\textwidth]{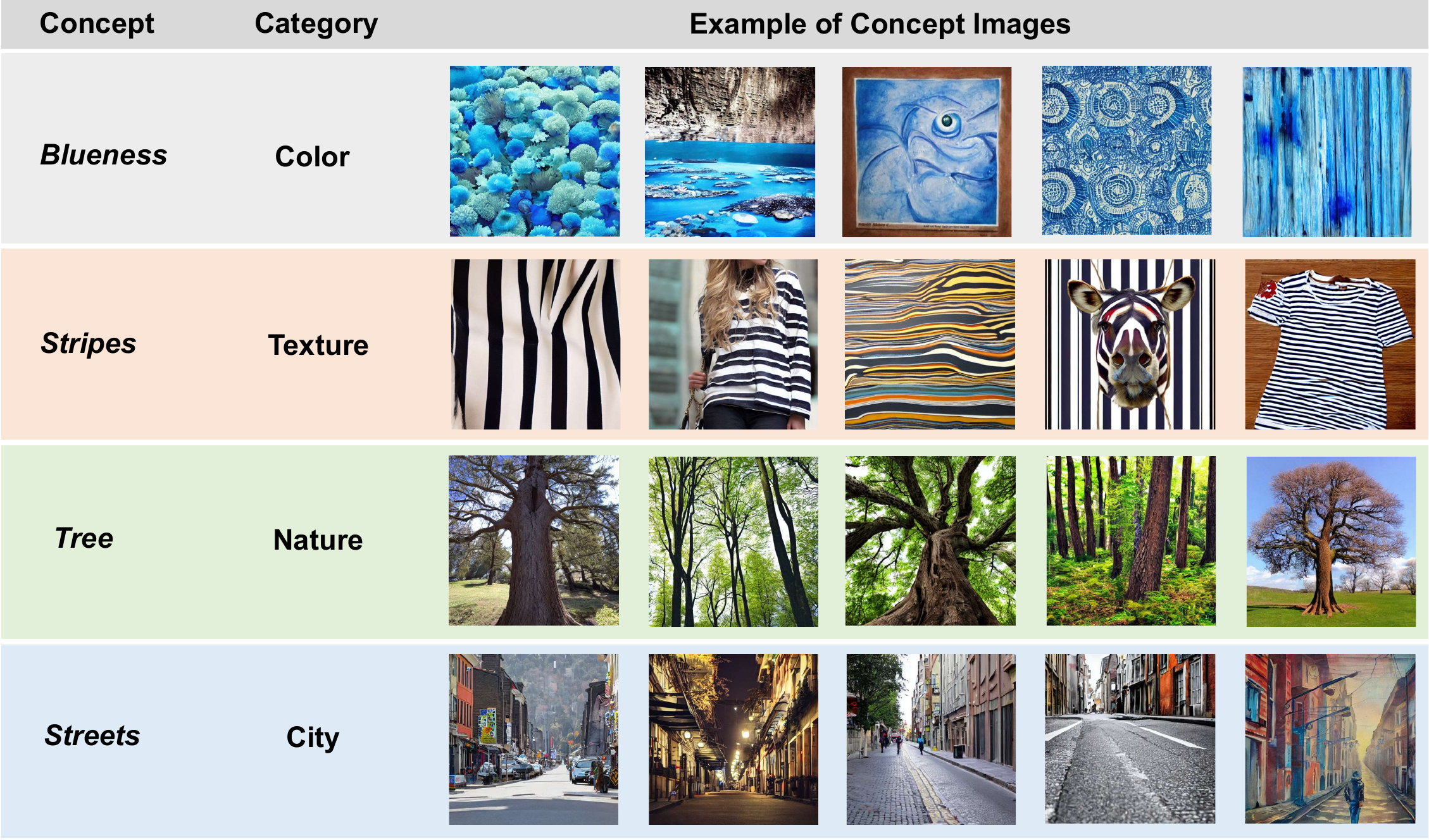}
    \caption{Examples of concept images in the concept bank.}
    \label{fig:demo}
\end{figure}

\xhdr{Concept image generation and examples} 
All the concept images are synthetic and generated by the Stable Diffusion model with the pretrained weights ``stable-diffusion-v1-4'', where we use the concept name or its pluralization form as prompts. The code of generating concept bank is made public at \href{https://github.com/Wuyxin/DISC/tree/master/concept_bank}{this link}. As shown in Figure~\ref{fig:demo}, we present the selected concept images in the concept bank as demonstrations. 

\vspace{5pt}

\xhdr{\rebuttal{Potential bias of concept images}}
Generative models may not necessarily be perfect at generating concept images. While it is true that they may have their own biases, they are trained on much larger datasets and thus are less likely to contain more severe spurious correlations for simple concepts. 
Empirically, we found that synthetic concept images are less noisy or biased compared to real images. For example, we observed that the concept images of “tree” in the BRODEN dataset~\cite{FongV18} of visual concepts highly coexist with “human” (\eg while hiking), while the synthetic images in our concept bank are much less likely to contain such bias, as shown in Figure~\ref{fig:demo}. 
Moreover, previous work~\cite{cce} also shows that learning the CAVs does not require a large number of concept images, which allows simple filtering on the concept bank to further guarantee its trustworthiness.

\begin{table}[H]
\centering
\caption{Selected concept categories for each dataset}
\label{tab:category}
\begin{tabular}{lcccccc}
\toprule
\multirow{2}{*}{}& \multicolumn{6}{c}{Concept categories}\\
& Color & Texture & Nature & City & Household & Others\\
\midrule
MetaShift & \textcolor{teal}{\CheckmarkBold}
& \textcolor{teal}{\CheckmarkBold} & \textcolor{teal}{\CheckmarkBold} & \textcolor{teal}{\CheckmarkBold} & \textcolor{teal}{\CheckmarkBold} & \textcolor{teal}{\CheckmarkBold} \\
Waterbirds & \textcolor{teal}{\CheckmarkBold} & \textcolor{teal}{\CheckmarkBold} & \textcolor{teal}{\CheckmarkBold}\\
FMoW & \textcolor{teal}{\CheckmarkBold} & \textcolor{teal}{\CheckmarkBold}& \textcolor{teal}{\CheckmarkBold} & \textcolor{teal}{\CheckmarkBold}\\
ISIC & \textcolor{teal}{\CheckmarkBold} & \textcolor{teal}{\CheckmarkBold}\\
\bottomrule
\end{tabular}

\end{table}
\xhdr{\rebuttal{Concept category selection and filtering}} As shown in Table~\ref{tab:category}, we select concept categories for each dataset. 
The general principle of selection is including the appeared objects in the dataset, based on the prior knowledge of the dataset context. We give the following demonstrations: 
\begin{itemize}[leftmargin=*]
    \item We include Color and Texture for all the datasets since these two concept categories have general existence.
    \item With the prior knowledge that FMoW is a satellite image dataset, we include Nature and City categories since they may appear in the dataset and thus could contain candidates of spurious concepts.
    \item With the prior knowledge that ISIC is a skin disease dataset, we exclude Nature, City, \etc, that do not exist from the concept candidates for this dataset, and only include Texture and Color concepts. 
\end{itemize}
 
In real-world applications, such knowledge of dataset contexts is fundamentally required for downstream tasks, which is \textbf{generalizable} to the other datasets. 
Moreover, since the large concept bank is shared across datasets and the practitioner can select the categories instead of the individual concepts, which requires \textbf{little labor}.

Moreover, in our implementation, we use a filtering module to filter relevant concepts in a dataset inspired by \citet{cce}.
The benefits of the concept category selection and filtering are (1) avoiding unrealistic interventions, \eg mixup animal images with satellite images, and (2) reducing the computational cost of computing CAVs during the training process.

\xhdr{Automatic concept category selection} 
As a future direction, to further avoid the concept category selection for an unknown downstream task, the protocol to automatically select suitable concept categories can be 

\begin{itemize}[leftmargin=*]
\item Leveraging image recognition models to identify existing objects in the datasets.
\item Then, extracting concepts or concept categories from our dataset-agnostic concept bank, which is defined in Table~\ref{tab:conceptbank}.
\end{itemize}

\xhdr{Learning CAVs} To learn the CAVs, we use $N^{p}=N^{n}=150$ for all the concepts. Another future direction is that we can learn more accurate concept representations by using hard negative samples. For example, we can construct the negative set for \textit{tree} concept using concepts images that are similar to tree images, \eg \textit{grass} and \textit{flowers}. For simplicity, we use random sampling to construct the negative sets in this work.

\section{Model and Optimization Details}
\label{app:models}
We adopt  DenseNet121~\cite{densenet} on FMoW and ResNet-50~\cite{resnet} on the other datasets. 
The hyper-parameters are summarized in Table~\ref{tab:hyper}. For the Beta distribution, we use $\alpha=\mu=2$ in all the datasets. Note that we search the number of clusters per class using Silhouette score, which is detailed in Appendix~\ref{app:ablation_sens}.

\begin{table}[H]
\centering
\caption{Hyper-parameters of \ours during training.}
\label{tab:hyper}
\begin{tabular}{lcccc}
\toprule
& Leaning Rate & Batch Size & Weight Decay & \#Clusters per Class\\
\midrule
MetaShift & $5e$-$4$ &$16$ &$1e$-$4$& $2$\\
Waterbirds & $1e$-$4$ & $32$ & $1e$-$4$& $3$\\
FMoW & $1e$-$4$& $10$& $0.0$& $3$\\
ISIC & $5e$-$4$& $16$ & $1e$-$5$& $3$\\
\bottomrule
\end{tabular}
\end{table}

\section{\rebuttal{Results of Interpretation Comparison}}
\label{app:interp}

Here we first analyze the advantages of the interpretations generated by DISC over the existing baselines that identify spurious correlation.
We study three dimensions of interpretability: 
\begin{itemize}
    \item \textbf{Class/group-wise}: Whether the explanations are concerning a class or group, which have the advantage of obtaining common insights across several instances, as opposed to instance-wise explanations.
    \item \textbf{Concept/caption-based}: Whether the explanations are based on captions or concepts that are more human-friendly and unambiguous instead of feature maps.
    \item \textbf{Adaptive}: Whether the explanations are adaptive or intrinsic during the training process, which enables dynamic inspection, as opposed to post-hoc explanations.
\end{itemize}

We consider different explanation types, including the existing saliency-based and concept-based methods. To highlight, DISC is the only method that fulfills the three advantages.

\begin{figure}[H]
    \centering \includegraphics[width=0.5\textwidth]{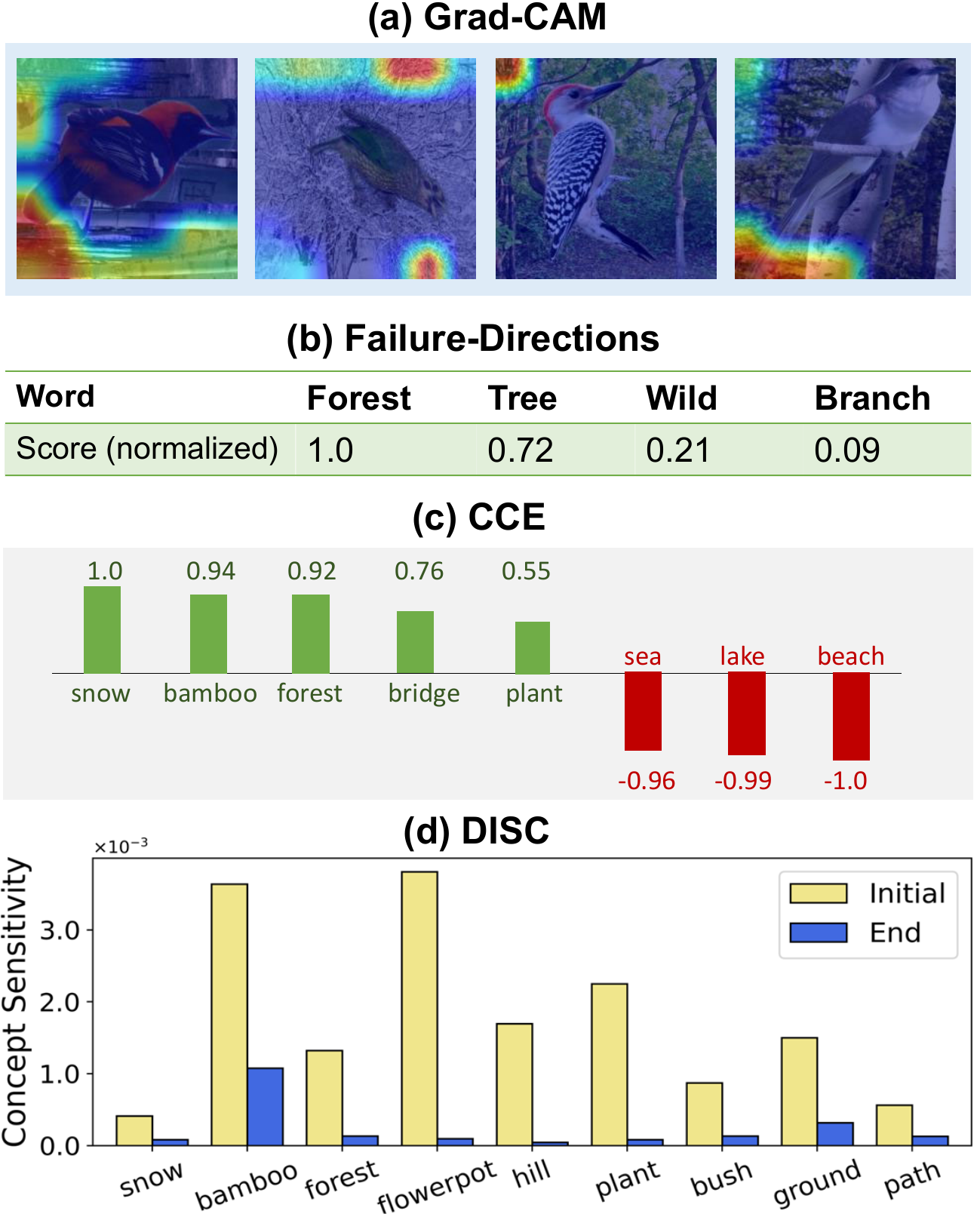}
    \vspace{-10pt}
    \caption{Different interpretations on Waterbirds explaining 
``\textbf{why the images are predicted as land birds}?'' . (a) Grad-CAM visualization. (b) The word score generated by \citet{distill-failure}.
    (c) The averaged concept scores when generating counterfactuals using CCE~\cite{cce}. (d) The concept sensitivity of spurious concepts on \textit{landbird} class before the after the \ours training.}
    \label{fig:interp}
\end{figure}

In Figure~\ref{fig:interp}, we further qualitatively evaluate the interpretations of \ours and three other types of explanations: (1) Grad-CAM (saliency-based method). (2) Failure-Direction~\cite{distill-failure} (caption-based method). (3) CCE~\cite{cce} (concept-based method). For Grad-CAM, similar to the previous observation, the instance-wise saliency maps could be hard to interpret and draw global insights in understanding the predictions for a class. For Failure-Direction, we compute the caption scores following the original paper and obtain the word scores by aggregation. Specifically, we found that the caption model sometimes focuses on the foreground instead of the background, making a subset of the captions uninformative for debugging.
Moreover, the interpretations lack diversity due to the limitation of the captioning model. For CCE, we find that the interpretations of \ours and CCE are similar. This aligns with our expectations since both \ours and CCE leverage CAVs to generate interpretations. Moreover, \ours offers more dynamic inspection during model training.

\begin{table}[H]
\centering
\caption{Comparison between interpretations of DISC and the existing methods.}
\label{tab:comparison}
\begin{tabular}{llccc}
\toprule
& Explanation types& Class/group-wise 	&Concept/caption-based	&Adaptive\\
\midrule
\citet{salient_imagenet}&\multirow{3}{*}{Saliency-based }& \multirow{3}{*}{\textcolor{teal}{\CheckmarkBold (partial)}}& \multirow{3}{*}{\textcolor{teal}{\CheckmarkBold (partial)}}&\multirow{3}{*}{\textcolor{teal}{\xmark}}\\
\citet{Grad-CAM}&\\
\citet{crm}& \\
\midrule
\citet{george}&\multirow{2}{*}{Clustering-based }& \multirow{2}{*}{\textcolor{teal}{\CheckmarkBold}}& \multirow{2}{*}{\textcolor{teal}{\xmark}}&\multirow{2}{*}{\textcolor{teal}{\xmark}}\\
\citet{bias-attribute}&\\
\midrule
\citet{eiil}& \multirow{4}{*}{Partition-based }&\multirow{4}{*}{\textcolor{teal}{\CheckmarkBold}}& \multirow{4}{*}{\textcolor{teal}{\xmark}}&\multirow{4}{*}{\textcolor{teal}{\xmark}}\\
\citet{jtt}& \\
\citet{DebiAN}& \\
\citet{sysgen}&\\
\midrule
\citet{cce}& \multirow{2}{*}{Concept-based}&\multirow{2}{*}{\textcolor{teal}{\CheckmarkBold}}& \multirow{2}{*}{\textcolor{teal}{\CheckmarkBold}}&\multirow{2}{*}{\textcolor{teal}{\xmark}}\\
\citet{partprot}\\
\midrule
\citet{distill-failure}& \multirow{2}{*}{Caption-based }& \multirow{2}{*}{\textcolor{teal}{\xmark}}& \multirow{2}{*}{\textcolor{teal}{\CheckmarkBold}}&\multirow{2}{*}{\textcolor{teal}{\xmark}}\\
\citet{Domino}& & &\\
\midrule
\citet{expstyle}&\multirow{2}{*}{Generative counterfactuals }& \textcolor{teal}{\xmark} &\textcolor{teal}{\xmark} &\textcolor{teal}{\xmark}\\
\citet{unknown_attribute}& &\textcolor{teal}{\CheckmarkBold}& \textcolor{teal}{\xmark}&\textcolor{teal}{\xmark}\\
\midrule
\textbf{\ourst}& Adaptive concept-based &\textcolor{teal}{\CheckmarkBold}& \textcolor{teal}{\CheckmarkBold}&\textcolor{teal}{\CheckmarkBold}\\
\bottomrule
\end{tabular}
\vspace{10pt}
\end{table}

\section{Results of Ablation and Sensitivity Study}
\label{app:ablation_sens}

\begin{table}[H]
    \vspace{-5pt}
    \caption{All experimental results of the ablation of model design choices.}
    \label{tab:all-ablation}
    \begin{center}
    \begin{tabular}{l|cc|cc|cc|c}
    \toprule
    \multirow{2}{*}{} & \multicolumn{2}{c|}{MetaShift} & \multicolumn{2}{c|}{Waterbirds} & \multicolumn{2}{c|}{FMoW} & {ISIC}\\
    & Avg. Acc.& Worst Acc.& Avg. Acc.& Worst Acc.& Avg. Acc.& Worst Acc.& Avg. AUROC \\\midrule
    \ourst-Randint       
    & 71.7\% 
    & 64.5\% 
    & 91.0\% 
    & 85.9\% 
    & 53.0\% %
    & 32.1\% %
    & 49.3\%
    \\
    \ourst-Reweight       
    & 72.8\% 
    & 62.5\%
    & 88.9\% 
    & 81.4\% 
    & 51.0\%
    & 32.0\%
    & 35.9\%
    \\
    \ourst-Inadaptive  &73.0\%	&68.3\%	&89.6\%	&86.5\%	&51.9\%	&31.8\%	&47.1\%\\
    \midrule
    \textbf{\ourst} 
    & 75.4\% 
    & 72.6\% 
    & 93.8\% 
    & 88.7\% 
    & 53.9\%
    & 36.1\%
    & 55.1\%
    \\
    \bottomrule
    \end{tabular}
    \vspace{-5pt}
    \end{center}
\end{table}

\xhdr{Ablation Results} In Table~\ref{tab:all-ablation}, we report the ablation results on all the datasets. The conclusions are consistent with our statements in the main paper. Specifically, \ours outperforms the ablation models by large margins, validating our algorithm design empirically.

\begin{figure}[H]
    \centering    \vspace{5pt}\includegraphics[width=1.0\textwidth]{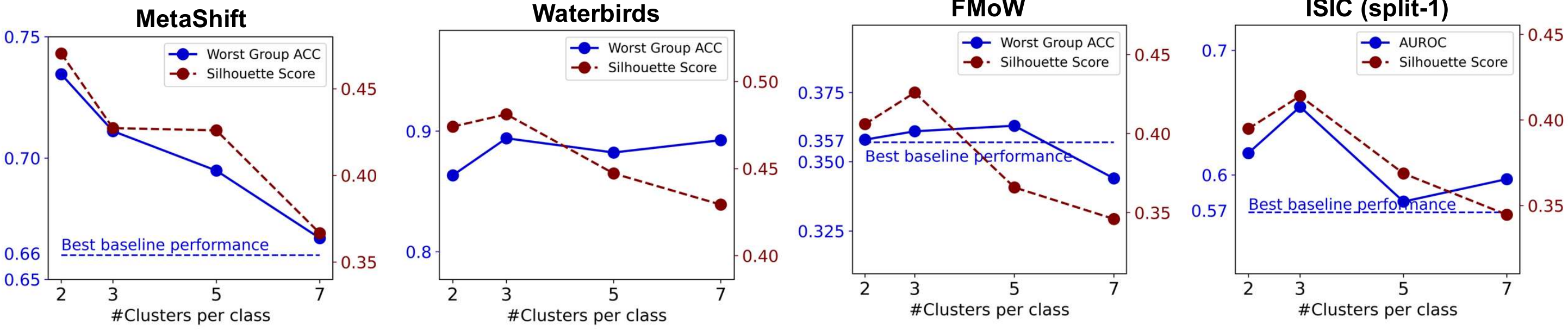}
    \caption{Worst Group Accuracy and Silhouette score \wrt number of clusters per class. For the ISIC dataset, we report the sensitivity result on one of the train-test splits.}
    \vspace{-5pt}
    \label{fig:all_sens}
\end{figure}

\xhdr{Unsupervised Clustering} We use the Silhouette score as a heuristic to search for the hyper-parameter of cluster number per class. As shown in Figure~\ref{fig:all_sens}, interestingly, we found this metric well aligns with the testing performances on most datasets. Empirically, we found a small number of clusters per class, \eg 3,  generally achieves the best results. One potential explanation is that when the number of clusters increases, the concept sensitivity could be passive and arbitrary by recognizing insignificant spurious concepts. We believe this is also an interesting perspective to investigate concept sensitivity or, in general, environment construction, in future works.

\begin{figure}[H]
    \centering
    \vspace{10pt}
    \includegraphics[width=0.8\textwidth]{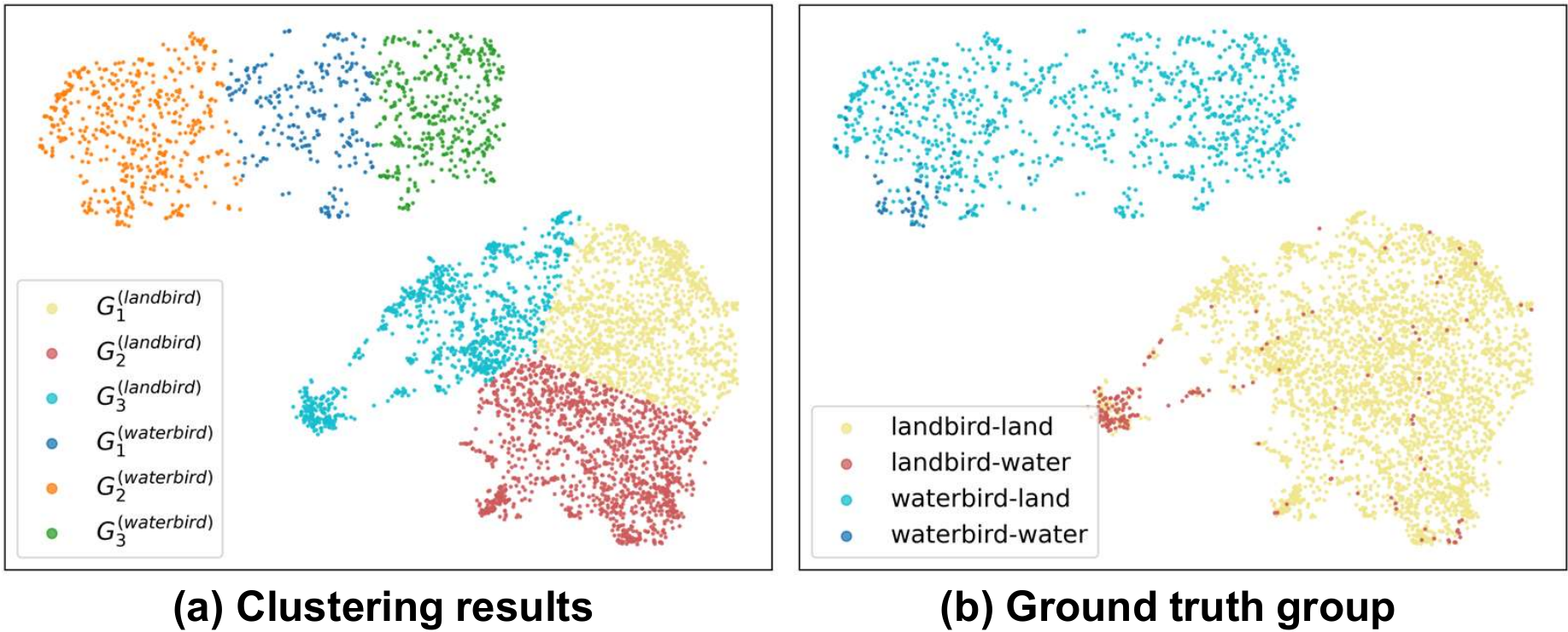}
    \vspace{-5pt}
    \caption{Comparison of clustering and group assignments on Waterbirds.}
    \label{fig:waterbirds_cluster}
\end{figure}

Besides the clustering results of MetaShift in Figure~\ref{fig:metashift}, we visualize the clustering results on Waterbirds in Figure~\ref{fig:waterbirds_cluster}. 
We found the clustering algorithm is able to capture part of the spurious attributes. Yet, good data environments could be difficult to find with extremely uneven groups. While \ours also outperforms most of the baselines, these results suggest that \ours is more robust even with ``imprecise'' environments.

\end{document}